\documentclass[preprint,12pt,3p]{elsarticle}

\journal{Information Sciences}

\usepackage{pythonhighlight}
\usepackage[utf8]{inputenc}
\usepackage[dvipsnames,table,xcdraw]{xcolor}
\usepackage{array}
\usepackage{textcomp}
\usepackage{stfloats}
\usepackage{url}
\usepackage{verbatim}
\usepackage{graphicx,tikz-cd}
\usepackage{subcaption}
\usepackage{lipsum}
\usepackage{enumitem}
\usepackage[colorlinks=true,linkcolor=blue,citecolor=blue,allcolors=blue]{hyperref}
\usepackage{algorithm}
\usepackage{algpseudocode}
\usepackage{arydshln}
\usepackage{adjustbox}
\usepackage{newfloat}
\usepackage{listings}
\usepackage{amsmath,amsfonts}
\usepackage{mathtools}
\usepackage{nicefrac}
\usepackage{amssymb}
\usepackage{amsthm}

\algblockdefx[Switch]{Switch}{EndSwitch}[1]{\textbf{switch} #1 \textbf{do}}{\textbf{end switch}}
\algblockdefx[Case]{Case}{EndCase}[1]{\textbf{case} #1:}{\textbf{end case}}
\algtext*{EndCase} 
\algblockdefx[Default]{Default}{EndDefault}{\textbf{default}:}{\textbf{end default}}
\algtext*{EndDefault} 

\usepackage{wrapfig}

\DeclareMathOperator*{\argmin}{arg\,min}

\definecolor{lb}{RGB}{31,119,180}

\newtheorem{theorem}{Theorem}[section]
\newtheorem{lemma}[theorem]{Lemma}
\newtheorem{proposition}[theorem]{Proposition}
\newtheorem{corollary}[theorem]{Corollary}

\theoremstyle{definition} 
\newtheorem{definition}{Definition}[section]
\newtheorem{example}{Example}[section]

\newtheorem{remark}{Remark}[section]

\usepackage{tcolorbox}
\newtcolorbox{mybox}[1]{colback=lb!1!white,colframe=lb!70!black,fonttitle=\bfseries,title=#1}

\usepackage{cleveref}

\begin{document}

\newcommand{\bruna}[1]{\textcolor{red}{\textbf{ [Bruna: #1] }}}

\begin{frontmatter}

    \title{Expert-Aided Causal Discovery of Ancestral Graphs}
    \author{%
        Tiago da Silva$^{8}$ \quad Bruna Bazaluk$^{6}$ \quad Eliezer de Souza da Silva$^{1,7}$ \quad António Góis$^2$ \\ \quad Salem Lahlou$^{8}$ \quad   Dominik Heider$^3$ \quad
        {Samuel Kaski}$^{4,5}$ \\ \quad  {Diego Mesquita}$^{1}$ \quad {Adèle Helena Ribeiro}$^{3}$ \\
        $^1$School of Applied Mathematics, Getulio Vargas Foundation; \\ $^2$Mila Quebec AI Institute, Universit\'e de Montr\'eal; \\
        $^3$Institute of Medical Informatics, University of Münster;\\
        $^4$Department of Computer Science, Aalto University; \\
        $^5$Department of Computer Science, University of Manchester; 
        \\$^6$Institute of Mathematics, Statistics and Computer Science, University of São Paulo; 
        \\$^7$Basque Center for Applied Mathematics; 
        \\$^8$Mohamed bin Zayed University of Artificial Intelligence.
        }
\def\figurewidth{0.5\textwidth}

\begin{abstract}
Causal discovery (CD) is an important component of many scientific applications, yet most techniques produce unreliable point estimates that often contradict expert knowledge.
To mitigate this, recent research has focused on \emph{ex-ante} incorporation of background knowledge into the CD process, typically under an unrealistic \emph{causal sufficiency} assumption.
When probing experts is costly (e.g., hidden behind expensive LLM APIs), however, ex-post model refinement that maximizes query utility is preferable.
Also, when independent experts provide conflicting but better-than-random feedback, a principled aggregation method is required.
In this context, we introduce the first CD algorithm that enables (i) distributional inference over ancestral graphs (AGs), which represent causal systems under latent confounding, and (ii) integration of both \emph{ex-ante} and uncertain \emph{ex-post} expert knowledge.
Briefly, our method is a diversity-seeking reinforcement learning algorithm, termed Ancestral GFlowNet (AGFN), whose policy we iteratively refine based on a Bayesian model of the noisy expert feedback.
Importantly, we prove convergence to the true AG given sufficiently accurate responses.
Through validation on synthetic and realistic datasets using simulated humans and LLMs, we show AGFN is competitive with or superior to strong baselines in terms of structural Hamming distance and Bayesian Information Criterion.
    \looseness=-1
\end{abstract}


\end{frontmatter}

\section{Introduction}

Causal discovery (CD) methods are widespread in science as tools for uncovering complex cause-and-effect relationships.
These algorithms typically leverage observational data to infer a graphical representation of the class of models that are equally likely to have generated the data, known as the Markov Equivalence Class (MEC).
In practice, however, they are known to be unreliable, as the statistical relationships drawn from the data may not align with the true causal model.
This mismatch defines what we call a violation of the \textit{faithfulness} assumption \citep{zhang2016faces}.
For instance, CD algorithms based on conditional independence tests may fall victim to false independence relations inferred due to a lack of statistical power.
Critically, these statistical errors may propagate and trigger a chain reaction of erroneous edge orientations \citep{Zhang2008, zhalama2017heuristic, ng2021reliable}.
Similarly, algorithms that maximize goodness-of-fit scores (e.g., BIC) may infer a structure that is optimal for the observed data, yet fails to properly represent either the ground-truth MEC~\citep{ogarrio2016gfci} or expert knowledge. \looseness=-1

\begin{wrapfigure}{l}{.28\textwidth}
    \vspace{-12pt}
    \includegraphics[width=\linewidth]{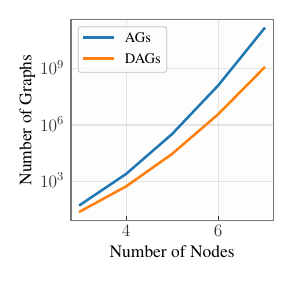}
    \caption{Number of AGs and DAGs per $\#$-nodes (log-scale).}
    \label{fig:nodesa}
    \vspace{-12pt}
\end{wrapfigure}
This issue becomes especially pronounced when unobserved, or latent, confounders are present. In such settings,
causal systems are commonly represented using \emph{ancestral graphs} (AGs),
which encode causal relationships while abstracting away latent confounding variables \cite{richardson2002ancestral}.
Specifically, in an AG, directed edges indicate that one variable is an ancestor of another---i.e., a cause through some directed path, which may involve intermediate or latent variables---and bidirected edges represent associations arising solely from latent confounding. Thus, a single AG can represent an equivalence class of causal diagrams that share the same ancestral structure.
When compared against the space of directed acyclic graphs (DAGs), which are commonly used to encode
\emph{causally sufficient}
systems---without latent confounders---, the space of AGs is significantly larger
(\Cref{fig:nodesa}).
Indeed, for six-variable datasets, we estimate there are approximately $1.3 \cdot 10^8$ possible AGs, roughly two orders of magnitude more than the $3.8 \cdot 10^6$ DAGs with the same number of nodes (see, e.g., \href{https://oeis.org/A003024}{OEIS A003024} and \Cref{sec:app:experimentsa} for our estimation method).
As such, recent research has largely focused on CD under causal sufficiency \cite{deleu2022bayesian, eitl_uai2025}.  \looseness=-1

In addition, while traditional CD algorithms such as FCI \cite{zhang2008fci}, GFCI \cite{ogarrio2016gfci}, ACI \cite{magliacane2016ancestral}, and DCD \cite{bhattacharya2021differentiable, ban2025differentiable} can handle latent confounding, they provide only a point estimate of the MEC, which may be insufficiently informative and potentially inconsistent with domain expertise.
Although recent studies have extended these techniques to accommodate background knowledge (BK) into their inferential processes (e.g., \cite{ribeiro2024anchorfci, DBLP:journals/ai/WangQZ23, DBLP:conf/aistats/Andrews20, eitl_uai2025}), these approaches primarily aim to ensure that the resulting graph satisfies a set of user-provided deterministic and noiseless constraints.
Additionally, by definition, BK can only be elicited in an \emph{ex-ante} fashion---prior to the algorithm's execution.
Arguably, however, an \emph{ex-post} refinement of the CD algorithm---whereby an expert can review high-likelihood graphs before providing feedback that refines the search---is no less critical in finding an accurate causal structure.
Moreover, when multiple experts provide possibly conflicting, but better-than-random, feedback, it is also important to update our search model according to such a valuable information.
This \emph{expert pooling} can only be achieved through an expert-in-the-loop (EITL) \cite{Amershi2014} pipeline that supports noisy feedback, which violates the premise of conventional approaches for EITL CD.
\looseness=-1

In this context, the central research questions (RQs) we investigate
are the following.

\begin{enumerate}[noitemsep, left=12pt, topsep=-2pt]
    \item[RQ1.] \emph{How to design a probabilistic CD algorithm under general latent confounding?}
    \item[RQ2.] \emph{How to ensure this algorithm is compatible with \emph{ex-ante} structural BK?}
    \item[RQ3.] \emph{How to refine the search model based on \emph{ex-post} and possibly noisy expert feedback?}
\end{enumerate}

To the best of our knowledge, our work is the first to jointly address these questions with an unified framework for probabilistic EITL CD under latent confounding.
We also emphasize that, by \emph{ex-ante} BK, we are speaking of \emph{rigid} structural constraints on the search space that are hard-coded into the CD algorithm prior to its training---e.g., sparsity can be enforced by constraining the nodes' degrees \cite{magliacane2016ancestral}.
Contrarily, \emph{ex-post} refinement refers to the iterative EITL procedure described earlier.
To achieve these objectives, we introduce Ancestral GFlowNets (henceforth, AGFN), a novel diversity-seeking reinforcement learning algorithm for CD under latent confounding.
Briefly, AGFN has two key ingredients: (i) an amortized sampler (GFlowNet; see \cite{bengio2021gflownet, bengio2021gflownetfoundations} and \Cref{sec:background}) that generates data-compatible AGs and (ii) an experimental design technique \cite{bharti22human} for interactively probing one or more experts on the relationship between a chosen pair of variables and updating the sampler according to their feedback.
\looseness=-1

\begin{figure}
    \centering
    \includegraphics[width=0.9\linewidth]{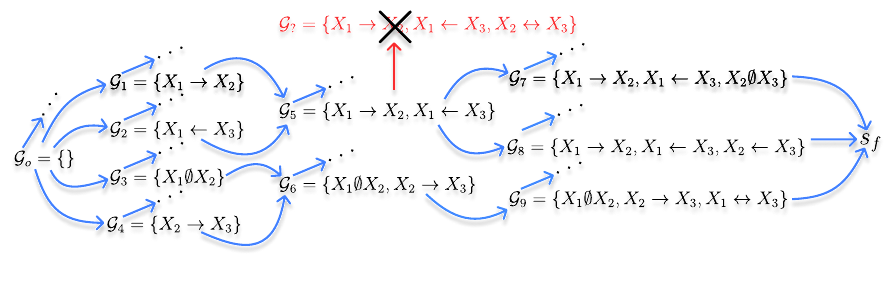}
    \caption{\textbf{Generative process for
    AGs.} We start at a (possibly edgeless) graph $\mathcal{G}_{o}$.
    At each state $\mathcal{G}_{t}$, we update the relationship between a pair of variables in $\mathcal{G}_{t}$ and proceed to the next state (\textcolor{blue}{blue}), until every pairwise relation has been specified.
    We mask out invalid actions resulting in graphs that are not ancestral (\textcolor{red}{red}). \looseness=-1
    }
    \vspace{-12pt}
    \label{fig:agfns}
\end{figure}

The design of our amortized sampler is illustrated in \Cref{fig:agfns}, where each graph is called a \emph{state}.
Simply put, we learn a stochastic policy function that, starting at a given initial AG, will select a pair of variables, define their relationship (either $\emptyset$, $\leftarrow$, $\rightarrow$, or $\leftrightarrow$), and proceed to the next state until every pairwise relation has been assigned.
We mask out actions leading to invalid (see Definition~\ref{def:ancestral}) states to ensure that each generated graph will be an AG.
In practice, we parameterize our policy function with a deep neural network, which is trained to sample each graph in proportion to a given goodness-of-fit score to the data \cite{deleu2024discreteprobabilisticinferencecontrol}.
In doing so, we ensure that the AGs that best fit the observed data are sampled most frequently.
Notably, unambiguous BK can be explicitly encoded into this edge-additive process.
For example, if it is known for certain that there is a bidirected edge between a given pair of variables, such an edge can be included in the initial AG.
Alternatively, if our interest lies solely on
sparse AGs, we can modify the policy function by zeroing out the probability of actions leading to a graph in which any of the nodes has a degree larger than a given threshold.
As we empirically show in \Cref{sec:experiments}, restricting the search for sparse solutions can also dramatically facilitate the sampler's the learning problem \cite{magliacane2016ancestral}. \looseness=-1

\begin{figure}
    \centering
    \includegraphics[width=0.6\linewidth]{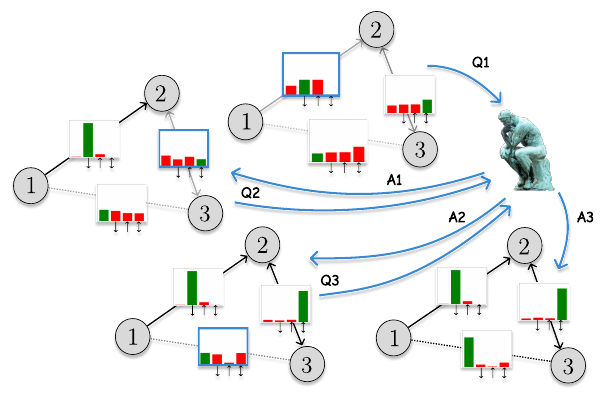}
    \caption{\textbf{EITL probabilistic CD} under latent confounding.
    We iteratively (i) probe an expert ($Q$) on the nature of the relationship between a chosen pair of variables (counterclockwise: $(1, 2)$, $(2, 3)$ and $(3, 1)$) and (ii) update our model based on the provided response ($A$).
    Histograms show marginals over edge types ($\emptyset$, $\rightarrow$, $\leftarrow$, $\leftrightarrow$) throughout this refinement process (\textcolor{ForestGreen}{green} denotes ground truth).
    Remarkably, our belief distribution over AGs becomes increasingly concentrated on the true AG, $1 \rightarrow 2 \leftrightarrow 3$, as shown in the bottom-right graph. \looseness=-1
    }
    \label{fig:eitla}
    \vspace{-12pt}
\end{figure}
Once an amortized sampler has been trained, we iteratively refine it by collecting noisy feedback from an expert. Throughout this work, we use the term \emph{expert} to denote an expert ensemble, which may consist of one or more human experts or a Large Language Model (LLM) queried multiple times
(see \Cref{sec:hitl}).
Importantly, we assume our expert provides a better-than-random assessment on the nature of the relationship (or lack thereof) between any pair of variables.
In \Cref{fig:agfns}, for example, suppose we asked an expert for the edge type between the nodes $X_{1}$ and $X_{2}$.
The expert responds with a single symbol in $\{\emptyset, \leftarrow, \rightarrow, \leftrightarrow\}$ indicating either the absence of a relationship ($\emptyset$), a directed edge from $X_{1}$ to $X_{2}$ ($\rightarrow$), a directed edge from $X_{2}$ to $X_{1}$ ($\leftarrow$), or a bidirected edge between $X_{1}$ and $X_{2}$ ($\leftrightarrow$).
This response is used to update a posterior belief distribution over the relationship between $X_{1}$ and $X_{2}$ via Bayes' rule, which we later
combine with the policy function via log-pooling \cite{robert2007bayesian}.
In this context, we implement our EITL as a two-step procedure.
First, we query the expert on the pair of variables that minimizes the expected entropy of the amortized sampler's distribution over AGs.
This approach is inherited from the literature of Bayesian experimental design \cite{ryan15bayesian} and active learning \cite{cohn1994active, DBLP:journals/ijar/HauserB14}, and we refer to it as \emph{active feedback elicitation}.
Second, we update our policy function based on the received feedback.
From this perspective, the proposed algorithm is inherently \emph{interactive}: expert feedback modifies the search model, which in turn requests further information from the expert until a stopping criterion is met.
This process progressively separates AGs that are both statistically plausible and factually consistent from those that are not within the amortized sampler's learned distribution.
Our EITL process in illustrated in \Cref{fig:eitla}. \looseness=-1

From this standpoint, AGFN addresses the three research questions 
posed above: it allows for probabilistic inference over the space of AGs (RQ1) and for the integration of both deterministic \emph{ex-ante} (RQ2) and noisy \emph{ex-post} (RQ3) expert knowledge into the inferential process.
As such, the remainder of the paper is organized as follows.
In \Cref{sec:background}, we provide an overview of AGs, latent confoundedness, and amortized sampling via entropy-regularized RL.
In \Cref{sec:ancestralGFlowNet}, we characterize our amortized sampler of AGs, and we demonstrate that the underlying distribution is supported on the space of AGs---i.e., all AGs are sampled with positive probability, while non-AGs are excluded from sampling.
In \Cref{sec:hitl}, we describe our framework for active knowledge elicitation and our Bayesian rule for updating the amortized sampler based on the expert's feedback. We also prove that the mode of the updated distribution over AGs converges to the true AG when the received feedback is sufficiently accurate.
In \Cref{sec:experiments}, we evaluate AGFN on synthetic and realistic datasets 
with both a simulated human and an LLM as experts, and we show that AGFN consistently outperforms strong baselines in terms of both the SHD to the true AG and the goodness-of-fit of the candidate AGs.
We also discuss related works in detail in \Cref{sec:app:relat_work}, and elaborate on our main findings, the limitations of our method, and directions for future research in \Cref{sec:conclusionsa}. 
In summary, our contributions are described below.

\begin{enumerate}[noitemsep, left=4pt, topsep=-2pt]
    \item We introduce 
    the first probabilistic CD method
    under general latent confounding. \looseness=-1
    \item We propose the first EITL pipeline supporting both
    structural constraints on the search space and noisy expert feedback for iterative refinement of the CD process. \looseness=-1
    \item We develop an optimal Bayesian experimental design to interactively query one or more experts on the most informative pair of variables at each stage of the EITL pipeline. \looseness=-1
    \item We empirically show that our method performs competitively with, and often significantly outperforms, strong baselines for CD under latent confounding,
    achieving high graph accuracy after receiving only a few (fewer than four) expert responses. \looseness=-1
\end{enumerate}


\section{Background}
\label{sec:background}

The causal reasoning used in our work is grounded in Pearl’s framework of causality \citep{pearl:2k}. We first introduce structural causal models, which provide a functional specification of the underlying causal system, including both observed and unobserved variables. We then introduce their associated graphical representations, known as causal diagrams, which convey a structural description of these models. On this basis, we motivate the use of AGs, which offer a more convenient and parsimonious representation of a causal structure while encoding more abstract forms of knowledge that align closely with the type of information typically available to researchers and domain experts (see a discussion in \Cref{sec:app:knownledge}). We conclude with a brief review of amortized sampling. The reader is invited to consult \cite{richardson2002ancestral, zhang2008causal} for a comprehensive exposition of causal reasoning with AGs, and \cite{deleu2024discreteprobabilisticinferencecontrol} for a clear introduction to amortized inference in discrete spaces. \looseness=-1

\paragraph{Notations} We begin with standard definitions.
We use possibly indexed uppercase letters (e.g., $U$, $V$, $V_i$, $V_j$) to denote random variables and bold uppercase letters (e.g., $\mathbf{U}$, $\mathbf{V}$) to represent sets or vectors of such variables.
We also represent an AG by $\mathcal{G}$ or $\mathcal{G}_{t}$ for some integer index $t$.
Unless stated otherwise, Greek letters will be reserved for random variables (e.g., $\omega$ and $\eta$), and boldface Greek letters will denote model parameters (e.g., $\boldsymbol{\theta}$).
As an exception, we let $\Delta_{m} = \{\mathbf{x} \in \mathbb{R}_{m} \colon \sum_{i=1}^{m} \mathbf{x}_{i} = 1 \text{ and } \mathbf{x} \ge 0\}$ be the $(m - 1)$-dimensional simplex. \looseness=-1

\begin{definition}[Structural Causal Model (SCM) \citep{pearl:2k}]
\label{def:scm}

An SCM $\mathcal{M}$ is defined as
a 4-tuple $\langle \mathbf{U}, \mathbf{V}, \mathcal{F}, P(\mathbf{U})\rangle$, where $\mathbf{U}$
represents the set of exogenous (unobserved) variables, and $\mathbf{V}$ denotes
the set of endogenous (observed) variables.
The collection $\mathcal{F} = \{f_i\}_{i=1}^{|\*{V}|}$ consists of functions
where each function $f_i \in \mathcal{F}$ describes how each endogenous
variable $V_i \in \*{V}$ is determined by its direct causes, which include
both its exogenous causes $\mathbf{U}_i \subseteq \mathbf{U}$ and its endogenous
causes (or parents) $Pa(V_i) \subseteq \mathbf{V} \setminus \{V_i\}$. Specifically,
    \begin{equation} \label{eq:scm}
        V_{i} = f_{i}(\mathrm{Pa}(V_{i}), \mathbf{U}_{i}).
    \end{equation}
Additionally, $P(\mathbf{U})$ represents the probability distribution over
the exogenous variables.
\looseness=-1
\end{definition}

An SCM is uniquely represented by a causal diagram, in which directed edges indicate
causal relationships, and dashed bidirected edges indicate latent confounding,
meaning the connected variables share an unobserved common cause.

\begin{definition}[Causal Diagram \citep{pearl:2k}]
\label{def:causal_diagram}
An SCM $\mathcal{M}$ induces a DAG augmented with bidirected edges---or an acyclic directed mixed graph (ADMG)---$G(\mathbf{V}, \mathbf{E}_G)$, commonly referred to as a \emph{causal diagram}, which encodes the structural relationships among $\mathbf{V} \cup \mathbf{U}$. Each vertex in $G$ corresponds to a variable $V_i \in \mathbf{V}$ and:
\begin{enumerate}[noitemsep, topsep=-2pt, left=4pt]
    \item For every $V_i \in \mathbf{V}$ and each parent $V_j \in \text{Pa}(V_i)$, there is a directed edge $V_j \to V_i \in \mathbf{E}_G$.
    \item For every pair $V_i, V_j \in \mathbf{V}$ sharing a common exogenous parent (i.e., $\mathbf{U}_i \cap \mathbf{U}_j \neq \emptyset$), there is a dashed bidirected edge $V_i \dashleftarrow \!\!\!\!\!\!\!\!\! \dashrightarrow V_j \in \mathbf{E}_G$ representing latent confounding.
\end{enumerate}
\looseness=-1
\end{definition}

Note that a causal diagram provides a purely qualitative description of the causal model, without specifying the functional forms or probability distributions of the variables. Thus, it represents an entire class of SCMs that share the same underlying causal structure.

\begin{figure*}
    \centering
    \includegraphics[width=.8\linewidth]{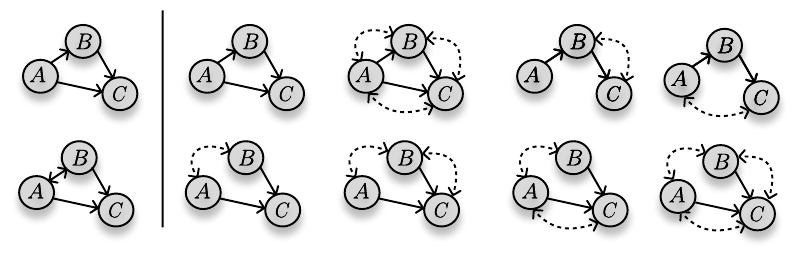}
    \caption{Two AGs (left panel) and a selection of compatible causal diagrams (right panel). Each row associates an AG together with some causal diagrams that are consistent with its implied ancestral relationships, illustrating how a single AG encodes a class of potentially many causal diagrams and corresponding SCMs.}
    \label{fig:ags_causaldiagrams}
\end{figure*}

\subsection{Ancestral graphs}

Consider a causal diagram $G$ representing the underlying SCM over $\mathbf{V}$. For a variable $V_i$, we define its set of ancestors as $\mathrm{An}(V_i) = \{V_j \in \mathbf{V} \colon V_j \rightsquigarrow V_i\}$, where $V_j \rightsquigarrow V_i$ denotes the existence of a directed (causal) path from $V_j$ to $V_i$ in $G$.

Assuming no selection bias, an \emph{ancestral graph} (AG) $\mathcal{G}$ is a \emph{mixed graph}, comprising both directed ($\rightarrow $) and bidirected ($\leftrightarrow$) edges, which preserves all ancestral relationships
in the underlying causal diagram $G$ \cite{richardson2002ancestral, DBLP:conf/uai/Zhang07}.
A directed edge ($V_i \rightarrow V_j$) in $\mathcal{G}$ indicates that $V_i \in \mathrm{An}(V_j)$. In contrast, a bidirected edge ($V_i \leftrightarrow V_j$) in $\mathcal{G}$ indicates that neither variable is an ancestor of the other, implying that the association between $V_i$ and $V_j$ arises solely from latent confounding.
Importantly, an AG is unambiguously defined only if it contains neither directed or almost directed cycles. We formalize this below.
In view of these additional structural constrains, sampling AGs is inherently more challenging than sampling DAGs.

\begin{definition}[Ancestral Graphs] \label{def:ancestral}
    Let $\mathcal{G} = (\mathbf{V}, \mathbf{E})$ be a mixed graph over vertices $\mathbf{V}$ and with edges $\mathbf{E}$, and let $\mathrm{An}(V_i)$ denote the ancestors of $V_i$.
    We say that $\mathcal{G}(\mathbf{V}, \mathbf{E})$ is \emph{ancestral} if, for every pair of vertices $V_{i}, V_{j} \in \mathbf{V}$, the following conditions hold: \looseness=-1
\begin{enumerate}[noitemsep, topsep=-2pt, left=4pt]
    \item \textbf{No directed cycles.} If $V_i \rightarrow V_j \in \mathbf{E}$, then $V_{j} \notin \mathrm{An}(V_{i})$. \looseness=-1
    \item \textbf{No almost directed cycles.}
    If $V_{i} \leftrightarrow V_{j} \in \mathbf{E}$, then $V_{i} \notin \mathrm{An}(V_{j})$ and $V_{j} \notin \mathrm{An}(V_{i})$. \looseness=-1
\end{enumerate}
\end{definition}

Notably, while directed edges in an AG encode ancestral relationships, they do not distinguish between direct and indirect causal paths and may reflect relations influenced by latent confounding.
Consequently, latent confounding may be present even between variables connected by a directed edge.
Different causal diagrams can therefore map to the same AG, making it a representation of an equivalence class of causal diagrams consistent with the implied ancestral relationships. This is illustrated in Figure \ref{fig:ags_causaldiagrams}, where the left panel shows two AGs and the right panel displays a selection of compatible causal diagrams.
As indicated, a directed edge in an AG (i.e., an implied ancestral relation) may arise from a causal path that is indirect, affected by latent confounding, or both.

\begin{remark}[On Markov Equivalent AGs]
Different AGs can entail the same set of conditional independencies
(for details, see \cite{DBLP:conf/uai/Zhang07}), as is the case for the two AGs shown in Figure~\ref{fig:ags_causaldiagrams}. Such AGs are referred to as \emph{Markov equivalent} and collectively form an \emph{MEC}. This implies that any SCM corresponding to a causal diagram compatible with these AGs generates the same observational distribution. Consequently, the likelihood of the observed data, and any likelihood-based score---such as BIC for linear Gaussian AGs \cite{richardson2002ancestral}, the Degenerate Gaussian (DG) score for linear AGs with mixed data types \cite{andrews19alearning}, and the multivariate information score for more complex AGs \cite{lagrange2025efficient}---, is identical for all SCMs in the MEC.
In practice, when searching for the true data generating AG, one typically considers the set of most likely or highest scoring AGs and treats as definitive those relationships that are shared across all of them. Importantly, when likelihoods are updated using expert knowledge, this set is further refined, yielding an equivalence class that is more informative than the MEC
by resolving some of the structural ambiguities that are indistinguishable from observational data alone.
\looseness=-1
\end{remark}

\subsection{Amortized Samplers} \label{sec:thisss}

We recall that our objective is to sample AGs from a given (user-specified) distribution.
As the space of AGs is discrete and combinatorial, we cannot rely on traditional gradient-based techniques (e.g., \cite{neal}).
Conventional approaches for categorical distributions, such as continuous relaxation and Gumbel softmax \cite{maddison2017concretedistributioncontinuousrelaxation, jang2017categoricalreparameterizationgumbelsoftmax}, also cannot be used due to the large number of AGs; see 
\Cref{fig:nodesa}.
Inspired by recent successes of Generative Flow Networks \citep[GFlowNets;][]{bengio2021gflownetfoundations, lahlou2023continuous} on amortized 
inference over combinatorial spaces and, in particular, on 
structure learning \cite{deleu2022bayesian, deleu2023joint, atanackovic2023dyngfn}, we propose
using an \emph{amortized sampler} over the space of AGs.
\looseness=-1

Before diving in, we first remind the reader of basic definitions and results on GFlowNets.
In a nutshell, a GFlowNet learns a \emph{policy function} on a \emph{state graph} by minimizing a stochastic objective enforcing a \emph{balance condition} (BC).
This BC, when satisfied, ensures that the marginal distribution of the policy over a set of \emph{terminal states} matches a given (perhaps unnormalized) distribution.
We make these terms precise below. \looseness=-1

\begin{definition}[State graph] \label{def:sgs}
    Let $\mathcal{S}$ be a finite set of states. We distinguish three types of specialized states: an \emph{initial} state $s_{o} \in \mathcal{S}$, a set of \emph{terminal} states $\mathcal{X} \subseteq \mathcal{S}$, and a \emph{final} state $s_{f} \in \mathcal{S}$.
    A \emph{state graph} (SG) over $\mathcal{S}$ is a DAG with the following properties.
    \begin{enumerate}[noitemsep, left=4pt]
        \item $s_{o}$ is the only state without incoming edges; 
        \item for each $s \in \mathcal{S}$, there is a path from $s_{o}$ to $s$;
        \item the only child of each $x \in \mathcal{X}$ is $s_{f}$; and
        \item $s_{f}$ is the only state without outgoing edges.
    \end{enumerate}
    In \Cref{fig:agfns}, for instance, $s_{o} = \mathcal{G}_{o}$, $\mathcal{G}_{i} \in \mathcal{S} \setminus \mathcal{X}$ for $i \in \{1, 2, \dots, 6\}$ and $\mathcal{G}_{j} \in \mathcal{X}$ for $j \in \{7, 8, 9\}$.
\end{definition}

We emphasize that a SG and an AG are completely distinct objects, and should not be confused.
Indeed, as in \Cref{fig:agfns}, an SG can have AGs as vertices.
That said, a GFlowNet is simply defined as a deterministic Markov decision process with a stochastic policy over a SG. \looseness=-1

\begin{definition}[GFlowNet] \label{def:gflownets}
    A \emph{GFlowNet} is a tuple $(\mathcal{S}, \mathcal{X}, s_{o}, s_{f},  \mathrm{SG}, p_{F}, p_{B})$ composed of a state graph $\mathrm{SG}$ over $\mathcal{S}$ with initial, terminal, and final states $s_{o}$, $\mathcal{X}$, and $s_{f}$, respectively, and \emph{forward} $p_{F} \colon \mathcal{S}^{2} \rightarrow [0, 1]$ and \emph{backward} $p_{B} \colon \mathcal{S}^{2} \rightarrow [0, 1]$ policies abiding by the following rules. \looseness=-1
    \begin{enumerate}[noitemsep, left=4pt]
        \item $p_{F}(s, \cdot)$ is a probability distribution over the children of $s$ in $\mathrm{SG}$ for $s \in \mathcal{S} \setminus \{s_{f}\}$.
        \item $p_{B}(s, \cdot)$ is a probability distribution over the parents of $s$ in $\mathrm{SG}$ for $s \in \mathcal{S} \setminus \{s_{o}\}$.
        \item $p_{F}(s_{f}, s_{f}) = 1$ and $p_{B}(s_{o}, s_{o}) = 1$, i.e., $s_{f}$ and $s_{o}$ are absorbing states for $p_{F}$ and $p_{B}$. 
    \end{enumerate}
\end{definition}

In practice, both $p_{F}$ and $p_{B}$ are parameterized as deep neural networks that receive a representation of $s \in \mathcal{S}$ as input and output a probability distribution over the children of $s$ in the SG.
Given a probability distribution $R \colon \mathcal{X} \rightarrow \mathbb{R}_{+}$, our objective is to learn a $p_{F}$ such that the marginal of $p_{F}$ over $\mathcal{X}$ matches $R$, i.e.,
\begin{equation} \label{eq:ma}
    p_{\intercal}(x) \coloneqq \sum_{\tau \colon s_{o} \rightsquigarrow x} \prod_{(s, s') \in \tau} p_{F}(s'|s) \coloneqq \sum_{\tau \colon s_{o} \rightsquigarrow x} p_{F}(\tau|s_{o}) \propto R(x),
\end{equation}
in which $s_{o} \rightsquigarrow x$ means that the trajectory starts at $s_{o}$ and finishes at $x$.
As a result of the terminological inheritance from the RL literature, we may also refer to $R$ as a \emph{reward function}.
Due to the intractability of the above sum, we instead search for a $p_{F}$ and $p_{B}$ satisfying a \emph{trajectory-wise} condition \cite{malkin2022trajectory}, avoiding explicit marginalization of the forward policy. \looseness=-1

\begin{proposition}[Trajectory Balance (TB) Condition \cite{malkin2022trajectory}] \label{prop:tba}
    Assume there is a constant $Z \in \mathbb{R}_{+}$ s.t. $Z p_{F}(\tau|s_{o}) = R(x)p_{B}(\tau|x)$ for each trajectory $\tau \colon s_{o} \rightsquigarrow x$ and each $x \in \mathcal{X}$. 
    Then $p_{\top}(x) \propto R(x)$ and $Z = \sum_{x \in \mathcal{X}} R(x)$, with $p_{\top}$ as in \Cref{eq:ma}. \looseness=-1
\end{proposition}

Towards this objective, we minimize an expectation of the log-squared difference between the left- and right-hand sides of the TB condition, $\mathbb{E}_{\tau \sim p_{\epsilon}}[(\log Z + \log p_{F}(\tau|s_{o}) - \log p_{B}(\tau|x) - \log R(x))^{2}]$, with respect to an $\epsilon$-greedy policy defined as $p_{\epsilon}(s, \cdot) = \epsilon p_{U}(s, \cdot) + (1 - \epsilon) p_{F}(s, \cdot)$ for $s \in \mathcal{S}$; $p_{U}$ is the uniform forward policy in SG \cite{malkin2022trajectory}.
Importantly, as the size of the support of $p_{F}(s, \cdot)$ is supposedly negligible when compared to $|\mathcal{S}|$\footnote{In \Cref{fig:nodesa}, for instance, each state has $\mathcal{O}(n^2)$ children, while the SG grows as $\mathcal{O}(2^n)$ for $n$-sized AGs. \looseness=-1}, learning a $p_{F}(s, \cdot)$ is far more computationally tractable than directly approximating a distribution over $\mathcal{X}$.

From a broader perspective, GFlowNets may be interpreted as (diversity-seeking) entropy-regularized RL \cite{deleu2024discreteprobabilisticinferencecontrol, tiapkin2024generativeflownetworksentropyregularized} or as amortized hierarchical variational samplers \cite{malkin2023gflownetsvariationalinference}.
As applicable, we adopt these perspectives to better contextualize our algorithm throughout the text. \looseness=-1


\section{Ancestral GFlowNets} \label{sec:ancestralGFlowNet}

This section describes the design and efficient implementation of AGFNs.
In \Cref{sec:agfns}, we define AGFN as a GFlowNet over AGs.
Then, in \Cref{sec:samplingags}, we introduce an efficient algorithm for incorporating structural BK into AGFN's generative process and ensuring that its forward policy is exclusively supported on AGs---assigning zero-probability to non-AGs. \looseness=-1

\subsection{AGFNs} \label{sec:agfns}

\paragraph{State Graph} Formally, an AGFN is a GFlowNet over a SG defined by the edge-additive process illustrated in \Cref{fig:agfns}.
The initial state is a given AG, $\mathcal{G}_{o}$, and $\mathcal{G} \rightarrow \mathcal{G}'$ if and only if $\mathcal{G}'$ differs from $\mathcal{G}$ by a single extra relationship, which can be vacuous (i.e., indicating no edge between the corresponding variable pair in the AG).
Clearly, for $n$ variables, each $\mathcal{G}$ is reachable from $\mathcal{G}_{o}$ in at most $n \cdot (n - 1) \cdot \nicefrac{1}{2}$ steps, and each $\mathcal{G}$ has at most $2 \cdot n \cdot (n - 1)$ children---four for each variable pair, corresponding to each relationship type ($\emptyset$, $\rightarrow$, $\leftarrow$, $\leftrightarrow$).
Towards rigorously characterizing AGFN, we first define an \emph{assignment function}. Henceforth, $\mathbf{V}^{(2)} = \binom{\mathbf{V}}{2}$\footnote{Recall that $\binom{\mathbf{V}}{2} \coloneqq \{\{V_{i}, V_{j}\} \colon V_{i}, V_{j} \in \mathbf{V} \text{ and } V_{i} \neq V_{j}\}$.}.
\looseness=-1

\begin{definition}[Assignment function]
    Let $\mathbf{V}$ be the set of endogenous variables, and $f \colon \mathbf{V}^{(2)} \rightarrow \{\emptyset, \rightarrow, \leftarrow, \leftrightarrow\}$ be an \emph{assignment function} defining, for each $\{V_{i}, V_{j}\} \subseteq \mathbf{V}$, the ancestral relationship between $V_{i}$ and $V_{j}$.
    For definiteness, relationships are lexicographically ordered, i.e., $f(\{V_{i}, V_{j}\}) = \rightarrow$ means $V_{i} \rightarrow V_{j}$ for $i < j$.
    Clearly, each AG can be uniquely represented by a $f$.
    For each 
    $S \subset \binom{\mathbf{V}}{2}$, the restriction $f|_S$ defines a \emph{partial assignment function}.
    \looseness=-1
\end{definition}

In this regard, our SG is simply defined by the set of functions $f$ whose image $f(\mathbf{V}^{(2)})$ characterizes an AG---and their restrictions to proper subsets of $\mathbf{V}^{(2)}$.
Under this interpretation, there is an edge from $\mathcal{G}$ to $\mathcal{G}'$ in the SG if their partial assignments, $f|_{S}$ and $f|_{S'}$, differ only by the restricting set and $S' = S \cup \{\{V_{i}, V_{j}\}\}$ for some variable pair $\{V_{i}, V_{j}\} \in \mathbf{V}^{(2)} \setminus S$.
For conciseness, we let $\mathcal{R} = \{\emptyset, \rightarrow, \leftarrow, \leftrightarrow\}$ be the set of potential relationships in an AG.
\looseness=-1

\begin{definition}[AGFN's SG] \label{def:agfnsga}
    Let $\mathcal{F} = \{f, f \colon \mathbf{V}^{(2)} \rightarrow \mathcal{R} \}$ be the set of assignment functions.
    For each $S \subseteq \mathbf{V}^{(2)}$, define $\mathcal{G}(f, S)$ as the graph induced by the restriction of $f$ to $S$.
    That is, $\mathcal{G}(f, S) = (\mathbf{V}, \mathbf{E}_{S, f})$, with $\mathbf{E}_{S, f} = \{ V_{i} \ f(\{V_{i}, V_{j}\}) \ V_{j} \colon f(\{V_{i}, V_{j}\}) \neq \emptyset \text{ and } \{V_{i}, V_{j}\} \subseteq S\}$.
    Then, let $S_{o} \subseteq \mathbf{V}^{(2)}$ and $f_{o} \colon S_{o} \rightarrow \mathcal{R}$ be an \emph{initial partial assignment}.
    In this setting,
    \begin{enumerate}[noitemsep, left=4pt]
        \item the state space $\mathcal{S}$ is $\mathcal{S} = \{\mathcal{G}(f, S) \colon f \in \mathcal{F}_{o} \text{ and } S \subseteq \mathbf{V}^{(2)}\}$, with $\mathcal{F}_{o} \coloneqq \{f \in \mathcal{F} \colon f|_{S_{o}} = f_{o} \text{ and } \mathcal{G}(f, \mathbf{V}^{(2)}) \text{ is an AG}\}$ as the set of assignment functions agreeing with $f_{o}$ on $S_{o}$; 
        \item $s_{o} = \mathcal{G}(f_{o}, S_{o})$ is our initial state; and 
        \item $\mathcal{X} = \{\mathcal{G}(f, \mathbf{V}^{(2)}) \colon f \in \mathcal{F}_{o}\}$.
    \end{enumerate}
    AGFN's SG is defined over $\mathcal{S}$ with initial state $s_{o}$, wherein there is an edge from $s$ to $s'$ in $\mathcal{S}$ if and only if there is an assignment function $f$ and subsets $S \subseteq S'$ of $\mathbf{V}^{(2)}$ for which $|S' \setminus S| = 1$ and $s = \mathcal{G}(f, S)$, $s' = \mathcal{G}(f, S')$.
    We illustrate this 
    for three variables (i.e., $|\mathbf{V}| = 3$) in \Cref{fig:nodesa}. \looseness=-1
\end{definition}

The initial partial assignment ($f_{o}$) in \Cref{def:agfnsga} allows us to encode structural BK---as we discuss in the next section.
Later, when describing our EITL pipeline for ex-post model refinement, we will notice that the noisy expert-provided feedback for a variable pair $\{V_{i}, V_{j}\}$ will be used to update a belief distribution over $f(\{V_{i}, V_{j}\}) \in \mathcal{R}$, which is the reason we model our SG as in \Cref{def:agfnsga} (differing from prior work on graph-structured amortized inference \cite{deleu2022bayesian, deleu2023joint}).
For concreteness, we provide an example of AGFN's SG for $|\mathbf{V}| = 2$. \looseness=-1

\begin{example}[AGFN's SG]
    In \Cref{def:agfnsga}, Let $S_{o} = \emptyset$ and $|\mathbf{V}| = 2$.
    Then, $\mathcal{X} = \{ \{V_{1} \rightarrow V_{2}\}, \{V_{1} \leftarrow V_{2}\}, \{V_{1} \leftrightarrow V_{2}\}, \{V_{1}\emptyset V_{2}\}\}$, $s_{o} = \{\}$, and $\mathcal{S} = \{s_{o}\} \cup \mathcal{X}$.
\end{example}

\begin{figure*}
    \centering
    \includegraphics[width=.86\linewidth]{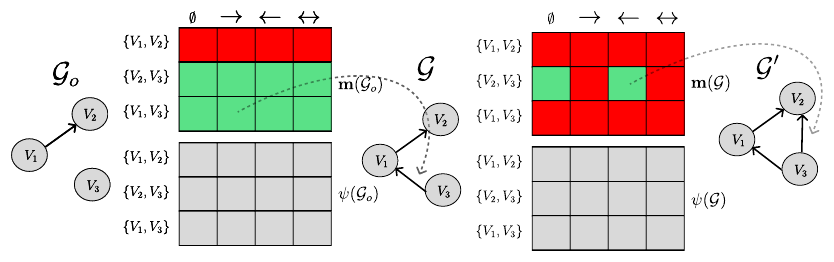}
    \caption{\textbf{AGFN's implementation} for variables $\mathbf{V} = \{V_{1}, V_{2}, V_{3}\}$. Each state $\mathcal{G}$ is associated with an estimated 
    distribution $\psi(\mathcal{G})$ and a 
    dynamic mask $\mathbf{m}(\mathcal{G})$ 
    defining the valid (\textcolor{green}{green}) and invalid (\textcolor{red}{red}) transitions at $\mathcal{G}$.
    We 
    sample a valid relationship (e.g., $V_{1} \leftarrow V_{3}$ in the first frame, and $V_{2} \leftarrow V_{3}$ in the second one) according to $\psi(\mathcal{G})$ and $\mathbf{m}(\mathcal{G})$, proceeding to the next state $\mathcal{G}'$ and updating the mask $\mathbf{m}(\mathcal{G}')$ as in \Cref{sec:samplingags}.
    }
    \label{fig:masking}
\end{figure*}

\paragraph{AGFN}
An AGFN is a GFlowNet built on top of the SG in \Cref{def:agfnsga}.
As such, we define the functions $q_{F} \colon \mathcal{S} \rightarrow \Delta_{4 \cdot \binom{n}{2}}$ and $q_{B} \colon \mathcal{S} \rightarrow \Delta_{\binom{n}{2}}$, which receive a state $s \in \mathcal{S}$ and output vectors $q_{F}(s)$ and $q_{B}(s)$ defining probability distributions over $\{1, \dots, 2 \cdot n \cdot (n - 1)\}$ and $\{1, \dots, \nicefrac{1}{2} \cdot n \cdot (n - 1)\}$, respectively.
Intuitively, $q_{F}(s)$ is a probability distribution over the ancestral relationships that can be \emph{added} to $s$, and $q_{B}(s)$ is a distribution over the ones that can be \emph{removed} from $s$.
Following the GFlowNet literature convention \cite{viviano2025torchgfnpytorchgflownetlibrary, deleu2022bayesian}, we let
\begin{equation*}
    q_{F}(s) = \mathrm{Softmax}( \psi(s) \odot \mathbf{m}(s)) \text{ and } q_{B}(s) \propto 1,
\end{equation*}
in which $\psi \colon \mathcal{S} \rightarrow \mathbb{R}^{4 \cdot \binom{n}{2}}$ is a (learnable) function, e.g., an MLP or a GNN \cite{kipf2017semisupervisedclassificationgraphconvolutional}, $\mathbf{m}(s) \in \{1, -\infty\}^{4 \cdot \binom{n}{2}}$ is a \emph{mask} ensuring that invalid transitions---either corresponding to already-defined relationships or resulting in non-AGs---are not realized by the policy function, and $\odot$ represents Hadamard's product.
We elaborate on the efficient computation of $\mathbf{m}(s)$ in the next subsection.
Given $q_{F}$ and $q_{B}$, which define categorical distributions, we simply pick each child (resp. parent) of $s$ accordingly to its index in $q_{F}(s)$ (resp. $q_{B}(s)$). 
Formally, we define indexing functions $\sigma_{s}^{\mathrm{Ch}} \colon \mathrm{Ch}(s) \rightarrow \{1, 2, \dots, 2 \cdot n \cdot (n - 1)\}$ and $\sigma_{s}^{\mathrm{Pa}} \colon \mathrm{Pa}(s) \rightarrow \{1, 2, \dots, 2 \cdot n \cdot (n - 1)\}$ of $\mathrm{Ch}(s)$ and $\mathrm{Pa}(s)$, respectively, with respect to $s$\footnote{
In \Cref{fig:masking}, for instance, $\sigma_{\mathcal{G}_{o}}^{\mathrm{Ch}}(\mathcal{G}) = 9$ and $\sigma_{\mathcal{G}}^{\mathrm{Ch}}(\mathcal{G}') = 6$, assuming row-major order for matrix storage.
}, and let \looseness=-1
\begin{equation} \label{eq:policiesa}
    p_{F}(s, s') = q_{F}(s)_{\sigma_{s}^{\mathrm{Ch}}(s')} \text{ and } p_{B}(s', s) = q_{B}(s)_{\sigma_{s}^{\mathrm{Pa}}(s')} \cdot \mathbf{1}_{\{s' \in \mathrm{Pa}(s)\}} 
\end{equation}
for each $s$ and $s'$ such that $s \rightarrow s'$ is a transition in the SG.
This is illustrated in \Cref{fig:masking}.

\subsection{Sampling AGs} \label{sec:samplingags}

\paragraph{Efficient mask computation}
Naively, we could compute $\mathbf{m}(\mathcal{G})$ at each state through the following procedure.
Let $\mathbf{A} = \mathbf{D} + \mathbf{B}$ be the adjacency matrix of an AG $\mathcal{G}$, and $\mathbf{D}$ and $\mathbf{B}$ be their directed and bidirected components.
Clearly, if $V_{i} \rightsquigarrow V_{j}$, we should mask both $V_{i} \leftrightarrow V_{j}$ and $V_{j} \rightarrow V_{i}$ in $\mathbf{m}(\mathcal{G})$.
Also, if there are variables $U, V$ for which $V_{i} \rightsquigarrow U$, $U \leftrightarrow V$, and $V \rightsquigarrow V_{j}$, we should also mask $V_{j} \rightarrow V_{i}$, as adding such an edge would result in an almost directed cycle.
These conditions can be verified in polynomial time via 
Floyd-Warshall's (FW) algorithm, which computes the pairwise (shortest-path) distances between nodes in a graph. \looseness=-1

\begin{lemma}[Algebraic form for (almost) directed paths] \label{lemma:cycles}
    Let $\mathcal{G} = (\mathbf{V}, E)$ be an AG, and $\mathbf{A} = \mathbf{D} + \mathbf{B}$ be $\mathcal{G}$'s decomposed adjacency matrix. 
    Define $\mathbf{F}$ as the pairwise distance matrix in $\mathcal{G}$ wrt $\mathbf{D}$ ($\mathbf{F}_{ij} = \infty$ if there is no path from $i$ to $j$) .
    Then, there is a directed path from $V_{i}$ to $V_{j}$ iff $\mathbf{F}_{ij} < \infty$, and there is an almost directed path from $V_{i}$ to $V_{j}$ iff $\min_{u, v} \mathbf{T}_{iuvj} < \infty$, with \looseness=-1
    \begin{equation*}
        \mathbf{T} = (\mathbf{F} \otimes \mathbf{1} \otimes \mathbf{1}) + (\mathbf{1} \otimes \tilde{\mathbf{B}} \otimes \mathbf{1}) + (\mathbf{1} \otimes \mathbf{1} \otimes \mathbf{F}) \in \mathbb{R}^{n \times n \times n \times n}, 
    \end{equation*}
    $\mathbf{1} \in \mathbb{R}^{n \times n}$ as the 
    matrix of $1$'s, and $\otimes$ is defined by $(\mathbf{A} \otimes \mathbf{B} \otimes \mathbf{C})_{iuvj} = \mathbf{A}_{iu} \cdot \mathbf{B}_{uv} \cdot \mathbf{C}_{vj}$ for 
    matrices $\mathbf{A}, \mathbf{B}, \mathbf{C}$ and indices $(i, u, v, j) \in \{1, \dots, n\}^{4}$, and $\tilde{\mathbf{B}}_{ij} = \mathbf{B}_{ij}$ if $\mathbf{B}_{ij} = 1$ else $\tilde{\mathbf{B}}_{ij} = \infty$. \looseness=-1
\end{lemma}

We provide the 
proof of Lemma~\ref{lemma:cycles} in \Cref{sec:app:pa}.
Based on this, we can define the mask $\mathbf{m}(\mathcal{G})$ as a simple concatenation of the boolean matrices $[\mathbf{F} < \infty]$ and $[\min_{u, v} \mathbf{T}_{iuvj} < \infty]$, wherein $[L] \coloneqq (1 \text{ if } L \text{ else } 0)$ represents the element-wise Inversion bracket for a clause $L$.
Although $\mathbf{F}$ and $\mathbf{T}$ can be efficiently computed in a GPU, repeatedly evaluating them during training can be expensive.
Instead, we propose an \emph{incremental} algorithm for iteratively updating the mask throughout the generative process.

For this, let $\boldsymbol{\alpha}(\mathcal{G}) \in \{1, 0\}^{n \times n}$ and $\boldsymbol{\beta}(\mathcal{G}) \in \{1, 0\}^{n \times n}$ be $n \times n$ matrices denoting, for each $(i, j)$, whether there is a directed path from $V_{i}$ to $V_{j}$ (i.e., $\boldsymbol{\alpha}_{ij}(\mathcal{G}) = 1$) and whether there is an \emph{almost} directed path from $V_{i}$ to $V_{j}$ ($\boldsymbol{\beta}_{ij}(\mathcal{G}) = 1$).
We compute (once) and store $\boldsymbol{\alpha}(\mathcal{G}_{o})$ and $\boldsymbol{\beta}(\mathcal{G}_{o})$ for the initial state $\mathcal{G}_{o}$ with Lemma~\ref{lemma:cycles}.
Then, upon the addition of a relation $V_{k} R V_{l}$ to $\mathcal{G}$, resulting in $\mathcal{G}'$, we update $\boldsymbol{\alpha}$ and $\boldsymbol{\beta}$ as in \Cref{alg:onlinea} ($R \in \{\emptyset, \rightarrow, \leftarrow, \leftrightarrow\}$).
Given $\mathcal{G}$, $\boldsymbol{\alpha}(\mathcal{G})$ and $\boldsymbol{\beta}(\mathcal{G})$, we can easily compute $\mathbf{m}(\mathcal{G})$ in \Cref{fig:masking}; see \textsc{UpdateRelationMask} in the pseudo-code. 
Crucially, both procedures in \Cref{alg:onlinea} can be efficiently implemented as linear operations with libraries such as \texttt{PyTorch} \cite{paszke2019pytorchimperativestylehighperformance} and \texttt{Jax} \cite{jax2018github}; we provide computer code for this in the supplement.
Additionally, as we show in the following proposition, this procedure ensures that AGFN's generative process is exclusively supported on the space of AGs.
\looseness=-1


\begin{algorithm}
\caption{Online algorithm for updating the mask of valid actions.}\label{alg:onlinea}
\begin{algorithmic}[1] 
\Require $V_{l}$, $V_{k}$ variables, relation $R \in \{\emptyset, \rightarrow, \leftarrow, \leftrightarrow\}$
\Require AG $\mathcal{G}$ and directed- and almost-directed path indicators $\boldsymbol{\alpha}, \boldsymbol{\beta} \in \{1, 0\}^{n \times n}$
\Require mask $\mathbf{m} \in \{1, -\infty\}^{4 \cdot \binom{n}{2}}$ at $\mathcal{G}$

\Procedure{UpdatePathMask}{$V_{k}, V_{l}, R, \mathcal{G}, \boldsymbol{\alpha}, \boldsymbol{\beta}$} 
    \State $\boldsymbol{\alpha}' \gets \boldsymbol{\alpha}$, $\boldsymbol{\beta}' \gets \boldsymbol{\beta}$

    \Switch{$R$}
        \Case{$\rightarrow$}
            \State $\boldsymbol{\alpha}'_{ij} \gets \max\{ \boldsymbol{\alpha}_{ij}, \boldsymbol{\alpha}_{ik} \cdot \boldsymbol{\alpha}_{lj}\}$ \text{ for } $(i, j) \in \{1, \dots, n\}^{2}$
            \State $\boldsymbol{\alpha}_{kl}' \gets 1$
            \State $\boldsymbol{\beta}'_{ij} \gets \max \{ \boldsymbol{\beta}_{ij}, \boldsymbol{\beta}_{ik} \cdot \boldsymbol{\alpha}_{lj}, \boldsymbol{\alpha}_{ik} \cdot \boldsymbol{\beta}_{lj}\}$ \text{ for } $(i, j) \in \{1, \dots, n\}^{2}$ 
        \EndCase
        \Case{$\leftarrow$}
            \State $\boldsymbol{\alpha}'_{ij} \gets \max\{ \boldsymbol{\alpha}_{ij}, \boldsymbol{\alpha}_{il} \cdot \boldsymbol{\alpha}_{kj}\}$ \text{ for } $(i, j) \in \{1, \dots, n\}^{2}$
            \State $\boldsymbol{\alpha}_{lk}' \gets 1$
            \State $\boldsymbol{\beta}'_{ij} \gets \max \{ \boldsymbol{\beta}_{ij}, \boldsymbol{\beta}_{il} \cdot \boldsymbol{\alpha}_{kj}, \boldsymbol{\alpha}_{il} \cdot \boldsymbol{\beta}_{kj}\}$ \text{ for } $(i, j) \in \{1, \dots, n\}^{2}$
        \EndCase
        \Case{$\leftrightarrow$}
            \State $\boldsymbol{\beta}_{ij}' \gets \max\{\boldsymbol{\beta}_{ij}, \boldsymbol{\alpha}_{ik} \cdot \boldsymbol{\alpha}_{lj}, \boldsymbol{\alpha}_{il} \cdot \boldsymbol{\alpha}_{kj} \}$ \text{ for } $(i, j) \in \{1, \dots, n\}^{2}$ 
        \EndCase
    \EndSwitch

    \State \Return $\boldsymbol{\alpha}', \boldsymbol{\beta}'$
\EndProcedure

\Procedure{UpdateRelationMask}{$V_{k}, V_{l}, R, \mathcal{G}, \boldsymbol{\alpha}, \boldsymbol{\beta}$} 
    \State $\mathbf{M} \gets \mathtt{reshape}\left(\mathbf{m}, \binom{n}{2} \times 4\right)$ \Comment{As in \Cref{fig:masking}: a row for each variable pair.}
    \State $\mathbf{M} \gets \mathtt{where}(\mathbf{M} > 0, 1, 0)$ \Comment{$\mathtt{reshape}$ and $\mathtt{where}$ represent 
    \texttt{NumPy} routines in \texttt{Python}.}
    \State $\boldsymbol{\alpha}', \boldsymbol{\beta}' \gets \textsc{UpdatePathMask}(V_{k}, V_{l}, R, \mathcal{G}, \boldsymbol{\alpha}, \boldsymbol{\beta})$
    \State $\mathbf{M}_{(k, l), (1, 2, 3 ,4)} \gets \mathbf{0} \in \mathbb{R}^{4}$ \Comment{Remark: we index the first axis of $\mathbf{M}$ with pairs $(k, l)$.}
    \For{$i = 1 \to n$, $j = i \to n$}
        \State $\mathbf{M}_{(i, j), 2} = 1 - \max\{\boldsymbol{\beta}_{ji}, \boldsymbol{\alpha}_{ji}\}$ \Comment{Mask $V_{i} \rightarrow V_{j}$ if $V_{j}$ and $V_{i}$ are path-connected.}
        \State $\mathbf{M}_{(i, j), 3} = 1 - \max\{\boldsymbol{\beta}_{ij}, \boldsymbol{\alpha}_{ij}\}$ \Comment{Same.}
        \State $\mathbf{M}_{(i, j), 4} = 1 - \max\{\boldsymbol{\alpha}_{ij}, \boldsymbol{\alpha}_{ij}\}$ \Comment{Mask $V_{i} \leftrightarrow V_{j}$ if $V_{j} \rightsquigarrow V_{i}$ or vice-versa.}
    \EndFor
    \State $\mathbf{m} \gets \mathtt{reshape}\left(\mathbf{M}, 4 \cdot \binom{n}{2}\right)$
    \Comment{Return transition and path masks for subsequent state.}
    \State \Return $\mathtt{where}\left(\mathbf{m} > 0, 1, -\infty\right), \boldsymbol{\alpha}', \boldsymbol{\beta}'$ 
\EndProcedure

\end{algorithmic}
\end{algorithm}

\begin{proposition}[AGFN only generates AGs] \label{prop:agfnsa}
    Assume $\mathbf{m}$ is updated as in \Cref{alg:onlinea}.
    Then, the generative process defined by the policy in \Cref{eq:policiesa} has support on the space of AGs abiding by the initial partial assignment function in \Cref{def:agfnsga}. \looseness=-1
\end{proposition}

Jointly, \Cref{def:agfnsga}, Lemma~\ref{lemma:cycles}, and Proposition~\ref{prop:agfnsa} completely characterize 
AGFN as an iterative generative process for AGs.
This forms the basis for our EITL pipeline for integrating expert knowledge into the CD process, which we describe next.

\paragraph{Introducing structural BK}
We start with the simplest form of knowledge, deterministic structural constraints. 
Under \Cref{def:agfnsga}, any information regarding the existence of a relationship between $V_{i}$ and $V_{j}$, e.g., $V_{i} \leftrightarrow V_{j}$, can be 
explicitly encoded into the partial assignment function $f_{o}$ defining the initial state.
As an illustration, $S_{o} = \{\{V_{i}, V_{j}\}\}$ and $f_{o}(\{V_{i}, V_{j}\}) = \leftrightarrow$ represents $V_{i} \leftrightarrow V_{j}$.
Similarly, we may enforce certain properties, such as sparsity, 
directly into \Cref{alg:onlinea}.
We collect a few examples below. \looseness=-1

\begin{example}[Sparsity]
    Often, we search for sparse causal structures \cite{magliacane2016ancestral}.
    In this case, let $\mathbf{i} \in \mathbb{R}^{n}$ and $\mathbf{o} \in \mathbb{R}^{n}$ be the maximum in- and out-degrees for the $n = |\mathbf{V}|$ nodes in $\mathcal{G}$.
    By modifying the \textsc{UpdateRelationMask} procedure in 
    \Cref{alg:onlinea}, we can easily keep track of each variable's degree and mask out transitions resulting in either in- or out-degrees larger than the ones in $\mathbf{i}$ and $\mathbf{o}$, respectively.
    We illustrate this in \Cref{sec:experiments}.
\end{example}

\begin{example}[Partitionability]
    In some settings, prior knowledge may indicate that causal relationships are more appropriately represented by partitioned graphs. This situation commonly arises in genotype–phenotype networks \cite{ribeiro2016causal}, or more generally when domain knowledge is more naturally specified at the level of clusters of variables, as formalized in the cluster-DAG (C-DAG) framework \cite{anand2023causal, anand2025cd}.
    In this case, we can 
    adjust the mask in \Cref{alg:onlinea} to ensure that directed edges are only added between nodes in different partitions. \looseness=-1
\end{example}

\begin{example}[Unconfoundedness]
    When we assume our dataset is causally sufficient, we may mask out all bidirected edges from AGFN's generative process.
    In doing so, we are restricting our search to the space of DAGs.
    From this perspective, our method subsumes prior work \cite{deleu2022bayesian, deleu2023joint} on structure learning 
    based on amortized inference. \looseness=-1
\end{example}


\section{EITL Causal Discovery}
\label{sec:hitl}

This section outlines our EITL framework for CD.
We recall that our objective is to update a trained AGFN (\emph{ex-post}) based on uncertain but better-than-random feedback regarding the ancestral relationship between a given pair of variables.
We formalize the concept of \emph{better-than-random} in Proposition~\ref{prop:consistencya};
in essence, it means that the feedback is more likely to be correct than to be wrong.
This situation might be encountered, for instance, when multiple experts provide independent feedback.
An LLM, viewed as an specialist, also produces independent and often self-conflicting responses to a query due to stochastic sampling.
Our approach enables expert-based iterative refinement of AGFN in both scenarios.
\looseness=-1

Simply put, then, our method works as follows.
As seen in \Cref{fig:masking}, the policy function $p_{F}(s, \cdot)$ induces a probability distribution over the relationships in $s$.
Based on a Categorical-Categorical model, the noisy feedback from an expert also characterizes a 
distribution $q_{\{V_{i}, V_{j}\}}(\cdot)$ over the relationships in a given (queried) pair of variables $\{V_{i}, V_{j}\}$.
By rewriting $p_{F}(s, \cdot)$ as a mixture distribution over relationship pairs, we combine 
$p_{F}(s, \cdot)$ and $q_{\{V_{i}, V_{j}\}}(\cdot)$ via log-pooling.
The intuition for this is that our resulting model, termed \emph{expert-refined AGFN}, should simultaneously assign high probability mass to both high-scoring ($p_{F}$) and expert-aligned ($q_{\{V_{i}, V_{j}\}}$) graphs, which can only be achieved by mixing them in log-space. 
Under certain conditions on $p_{F}$ and $q_{\{V_{i}, V_{j}\}}$, we show that the mode of the learned generative converges to the true AG almost surely as the number of received feedback increases. \looseness=-1

\subsection{A Bayesian model for the expert}

Our Bayesian model for the expert assumes that their feedback is a noisy realization of the true (unobserved) relationship between the queried pair of variables.
That is, let $r = \{V_{i}, V_{j}\}$.
Also, define $\boldsymbol{\rho}_{r} \in \Delta_{4}$ as a prior distribution over $(\emptyset, \rightarrow, \leftarrow, \leftrightarrow)$.
As our focus lies on ex-post refinement of AGFN, $\boldsymbol{\rho}_{r}$ may be informed by our trained amortized sampler.

As such, we define our model as follows.
Let $\pi_{r} \in [0, 1]$ be the (assumed fix) confidence of our expert.
Similarly, for $\omega \in \{\emptyset, \rightarrow, \leftarrow, \leftrightarrow\}$, let $\boldsymbol{\delta}_{\omega}$ be a Dirac function at $\omega$ (e.g., $\boldsymbol{\delta}_{\emptyset} = (1, 0, 0, 0)$ and $\boldsymbol{\delta}_{\rightarrow} = (0, 1, 0, 0)$, etc.) and $\mathbf{1} = (1, 1, 1, 1)$.
Then, the expert's feedback $f_{r}$ is assumed to be drawn according to the following 
hierarchical Bayesian model.
\begin{equation} \label{eq:bayesian}
    \begin{aligned}
        \omega_{r} &\sim \mathrm{Cat}\left( \boldsymbol{\rho}_{r} \right) \\
        f_{r} | \omega_{r} &\sim \mathrm{Cat} \left ( \pi_{r} \cdot \boldsymbol{\delta}_{\omega} + \left( \frac{1 - \pi_{r}}{3} \right) \cdot (\mathbf{1} - \boldsymbol{\delta}_{\omega}) \right).
    \end{aligned}
\end{equation}
Intuitively, $\omega_{r}$ represents the true ancestral relationship between $V_{i}$ and $V_{j}$.
In this regard, the distribution for $f_{r} | \omega_{r}$ means that the expert provides the correct response with frequency $\pi_{r}$, otherwise, their feedback is uniformly picked among the wrong alternatives with probability $1 - \pi_{r}$.
When $\pi_{r}$ cannot be confidently elicited for each variable pair $r$, we can either assume that the $f_{r}$'s are not identically distributed, with each query having a different $\pi_{r}$ (e.g., an LLM's verbalized confidence \cite{zhang2024calibratingconfidencelargelanguage, yang2024verbalizedconfidencescoresllms}), or put a prior distribution on $\pi_{r}$.
\looseness=-1

Under these circumstances, we may use a Beta-Categorical model for estimating the expert's confidence, $\pi_{r}$, based on their feedback.
For this, \Cref{eq:bayesian} can be modified to \looseness=-1
\begin{equation} \label{eq:bayesianpriors}
    \begin{aligned}
        \pi_{r} &\sim \mathrm{Beta}(\alpha_{r}, \beta_{r}) \\
        \omega_{r} &\sim \mathrm{Cat}\left( \boldsymbol{\rho}_{r} \right) \\
        f_{r} | \omega_{r}, \pi_{r} &\sim \mathrm{Cat} \left ( \pi_{r} \cdot \boldsymbol{\delta}_{\omega} + \left( \frac{1 - \pi_{r}}{3} \right) \cdot (\mathbf{1} - \boldsymbol{\delta}_{\omega}) \right), 
    \end{aligned}
\end{equation}
in which $\alpha_{r} > 0$ and $\beta_{r} > 0$ are hyperparameters.
Clearly, both \Cref{eq:bayesian,eq:bayesianpriors} result in a semi-tractable posterior distribution for $\omega_{r}$ with a simple sufficient statistic. \looseness=-1 

\begin{lemma}[Posterior distribution] \label{lemma:posteriora}
    Let $\mathbf{f}_{r}^{N} = (f_{r}^{(1)}, \dots, f_{r}^{(N)}) \in \{\emptyset, \rightarrow, \leftarrow, \leftrightarrow\}^{N}$, with $N$ as the number of feedback instances for $r$.
    The posterior 
    of $\omega_{r}$ 
    wrt $\mathbf{f}_{r}^{N}$ under \Cref{eq:bayesian} is \looseness=-1
    \begin{equation*}
        q_{r}(\omega_{r} | \mathbf{f}_{r}^{N}) \propto \boldsymbol{\rho}_{r}^{(\omega)} \cdot \pi^{n_{\omega_{r}}} \cdot \left( \frac{1 - \pi}{3} \right)^{N - n_{\omega_{r}}},
    \end{equation*}
    in which $n_{\omega_{r}} = \sum_{n=1}^{N} [f_{r}^{(n)} = \omega_{r}]$.
    Similarly, the posterior 
    of $\omega_{r}$ under \Cref{eq:bayesianpriors} is
    \begin{equation*}
        q_{r}(\omega_{r} | \mathbf{f}_{r}^{N}) \propto 3^{n_{\omega_{r}} - N} B(n_{\omega_{r}} + \alpha, N - n_{\omega_{r}} + \beta),
    \end{equation*}
    in which $B(x, y) \coloneqq \nicefrac{\Gamma(x)\Gamma(y)}{\Gamma(x + y)}$ is the Beta function and $x \mapsto \Gamma(x)$, the Gamma function.
\end{lemma}

Remarkably, when the $f_{r}$'s are drawn from either \Cref{eq:bayesian} or~\eqref{eq:bayesianpriors} conditioned on ground-truth parameters $\pi_{r}^{\star}$ and $\omega_{r}^{\star}$, and $\pi^{\star}$ 
represents a \emph{better-than-random} confidence (i.e., $\pi^{\star} > \frac{1}{4}$), 
our posterior distribution over $\omega_{r}$ converges to a delta Dirac at $\omega_{r}^{\star}$.
In other words, our model is \emph{consistent}.
We make this rigorous in the following proposition.

\begin{figure*}
    \centering
    \includegraphics[width=.9\linewidth]{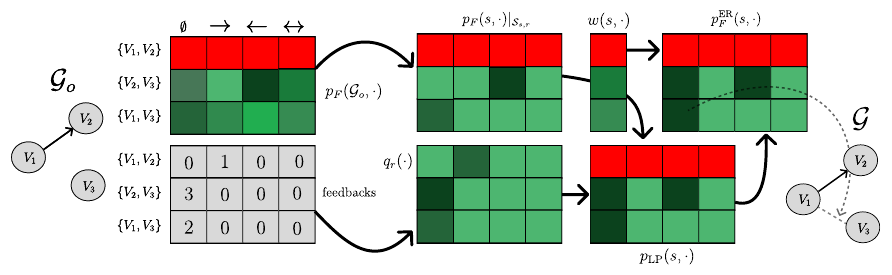}
    \caption{\textbf{Incorporating expert knowledge into AGFN}.
    Given 
    a forward policy $p_{F}$ (top, left) and expert responses (measured as counts; bottom, left), we normalize $p_{F}$ (middle, top) and compute the expert 
    posterior $q_{r}$ (middle, bottom) per variable pair.
    Then, we compute the log-pool of both distributions $p_{\mathrm{LP}}$ (bottom, right), which is used for selecting the next state (top, right), shown in the right-most panel. \looseness=-1
    }
    \label{fig:eitlexperta}
\end{figure*}

\begin{wrapfigure}{l}{.3\textwidth}
    \vspace{-8pt}
    \includegraphics[width=\linewidth]{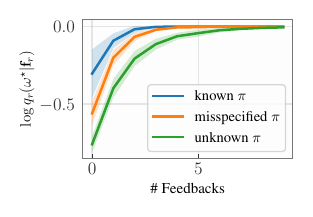}
    \caption{The posterior contracts around the true ancestral relationship ($\omega^{\star}$).}
    \label{fig:posteriora}
    \vspace{-12pt}
\end{wrapfigure}
\begin{proposition}[Consistency] \label{prop:consistencya}
    Assume $\mathbf{f}_{r}^{N}$ in Lemma~\ref{lemma:posteriora} is drawn from \Cref{eq:bayesianpriors} for certain parameters $\pi_{r}^{\star}$ and $\omega_{r}^{\star}$, in which $\omega_{r}^{\star}$ corresponds to the true relationship between the variables in $r$.
    Also, let $\pi_{r}^{\star} > \nicefrac{1}{4}$, i.e., the expert's feedback is \emph{better-than-random}.
    Then,
    \begin{equation*}
        q_{r}(\omega_{r}^{\star} | \mathbf{f}_{r}^{N}) \rightarrow 1
    \end{equation*}
    almost surely as $N \rightarrow \infty$ for both posteriors in Lemma~\ref{lemma:posteriora}.
    We illustrate this in \Cref{fig:posteriora}.
\end{proposition}

Remarkably, our model in \Cref{eq:bayesian} is also consistent under mild misspecification, i.e., when $\pi_{r}$ is incorrectly defined.
We show this in the next proposition.

\begin{proposition}[Consistency under misspecification] \label{prop:mconsistencya}
    Under 
    Proposition~\ref{prop:consistencya}, consider the model in \Cref{eq:bayesian} with a confidence parameter $\pi_{r} \neq \pi_{r}^{\star}$. That is, our Bayesian scheme is \emph{misspecified}.
    Nonetheless, if $\pi_{r} > \nicefrac{1}{4}$ and the expert's feedback is better-than-random,
    \begin{equation*}
        q_{r}(\omega_{r}^{\star} | \mathbf{f}_{r}^{N}) \rightarrow 1 \text{ almost surely as } N \rightarrow \infty.
    \end{equation*}
\end{proposition}

Given a posterior $q_{r}(\omega_{r} | \mathbf{f}_{r}^{N})$ over the relationship $\omega_{r} \in \{\emptyset, \rightarrow, \leftarrow, \leftrightarrow\}$ of the variable pair $r$, how do we combine $q_{r}$ with our 
sampler $p_{F}$? We elaborate on this in the next 
section.
\looseness=-1

\subsection{Belief updates}

To understand the connection between the posterior $q_{r}(\cdot)$ and 
AGFN's policy $p_{F}(s, \cdot)$, notice that $p_{F}(s, \cdot)$ may be seen as a distribution over the product space $\mathbf{V}^{(2)} \times \{\emptyset, \rightarrow, \leftarrow, \leftrightarrow\}$, in which $\mathbf{V}^{(2)}$ refers to the set of (unordered) variable pairs.
(We will henceforth omit the 
dataset $\mathbf{f}_{r}$ from $q_{r}$ for clarity).
By defining $t(s, \{V_{i}, V_{j}\}, c)$ as the transition function resulting from attaching the relationship $c \in \{\emptyset, \rightarrow, \leftarrow, \leftrightarrow\}$ to the pair $\{V_{i}, V_{j}\}$ and $\mathcal{S}_{s, \{V_{i}, V_{j}\}} = \{t(s, \{V_{i}, V_{j}\}, c) \colon c \in \{\emptyset, \rightarrow, \leftarrow, \leftrightarrow\}\}$ as the corresponding set of transitionable states from $s$ by modifying $\{V_{i}, V_{j}\}$,
we may rewrite $p_{F}(s, \cdot)$ as a mixture of the form \looseness=-1
\begin{equation} \label{eq:mixturesa}
    p_{F}(s, \cdot) = \sum_{\{V_{i}, V_{j}\} \in \mathbf{V}^{(2)}} w(s, \{V_{1}, V_{2}\}) \sum_{c \in \{\emptyset, \rightarrow, \leftarrow, \leftrightarrow\}} \delta_{t(s, \{V_{i}, V_{j}\}, c)}(\cdot) p_{F}(s, \cdot)|_{\mathcal{S}_{s, \{V_{i}, V_{j}\}}},
\end{equation}
in which $\delta_{t(s, \{V_{i}, V_{j}\}, c)}$ is a Dirac delta function, $p_{F}(s, \cdot)|_{\mathcal{S}_{s, \{V_{i}, V_{j}\}}}$ is the restriction of the policy to $\mathcal{S}_{s, \{V_{i}, V_{j}\}}$, and $w \colon \mathcal{S} \times \mathbf{V}^{(2)} \rightarrow [0, 1]$ is a weighting function defined by
\begin{equation*}
    w(s, \{V_{i}, V_{j}\}) \coloneqq \sum_{c \in \{\emptyset, \leftarrow, \rightarrow, \leftrightarrow\}} p_{F}(s, t(s, \{V_{i}, V_{j}\}, c)).
\end{equation*}
Clearly, $\sum_{r \in \mathbf{V}^{(2)}} w(s, r) = 1$ if $r$ has not yet been specified in $s$. In simpler terms, we can select a novel state $s'$ according to $p_{F}(s, \cdot)$ by first choosing a variable pair $\{V_{i}, V_{j}\}$ based on $w(s, \cdot)$ and then picking a $c \in \{\emptyset, \leftarrow, \rightarrow, \leftrightarrow\}$ from the restricted policy in \Cref{eq:mixturesa}, which can be seen as a distribution over $\{\emptyset, \leftarrow, \rightarrow, \leftrightarrow\}$.
We illustrate this in \Cref{fig:eitlexperta}.
From this perspective, our algorithm for \emph{belief updating} combines both $p_{F}(s, \cdot)|_{\mathcal{S}_{s, \{V_{i}, V_{j}\}}}$ and the expert-induced posterior over $r = \{V_{i}, V_{j}\}$ ($q_{r}(\cdot)$) via \emph{log-pooling}; see the next definition.

\begin{definition}[Expert-based policy refinement] \label{def:experta}
    For each $r \in \mathbf{V}^{(2)}$ and $s \in \mathcal{S}$, let $\mathcal{S}_{s, r}$ be as in \Cref{eq:mixturesa}.
    Also, let $t(s, r, c)$ be the state resulting from adding the ancestral relationship $c \in \{\emptyset, \leftarrow, \rightarrow, \leftrightarrow\}$ to $r$.
    Then, given a posterior $q_{r}$ in Lemma~\ref{lemma:posteriora} and a policy $p_{F}(s, \cdot)$ represented as in \Cref{eq:mixturesa}, we define the \emph{log-pooled} policy as
    \begin{equation*}
        p_{\mathrm{LP}}(s, (r, c)) \propto p_{F}(s, t(s, r, c))|_{\mathcal{S}_{s, r}}^{\beta_{N_{r}}^{(r)}} \cdot q_{r}(c)^{1 - \beta_{N_{r}}^{(r)}},
    \end{equation*}
    in which $\beta_{N_{r}}^{(r)} \in [0, 1]$ is a $r$-dependent schedule non-increasing on the number $N_{r}$ of feedback instances received so far. By replacing the right-hand side of~\eqref{eq:mixturesa} with its log-pooled counterpart, we obtain \looseness=-1
    \begin{equation*}
        p_{F}^{\mathrm{ER}}(s, \cdot) = \sum_{\{V_{i}, V_{j}\} \in \mathbf{V}^{(2)}} w(s, \{V_{1}, V_{2}\}) \sum_{c \in \{\emptyset, \rightarrow, \leftarrow, \leftrightarrow\}} \delta_{t(s, \{V_{i}, V_{j}\}, c)}(\cdot) p_{\mathrm{LP}}(s, (\{V_{i}, V_{j}\}, c)). 
    \end{equation*}
    This characterizes the \emph{expert-refined} (ER) policy function for our model.
\end{definition}
Instructively, notice that our ER policy reduces to a tempered version of the AGFN's policy when no feedback has been provided (so $q_{r}(\cdot)$ becomes the prior) 
and our prior distribution 
in \Cref{eq:bayesian} is uniform.
A corollary of Proposition~\ref{prop:consistencya} is that, given a 
sufficient number of better-than-random feedback responses, the unique mode of the AGFN's distribution will be the true AG.
Asymptotically, AGFN converges to a Dirac delta on this graph. \looseness=-1
\begin{corollary}[AGFN is consistent] \label{col:consistencya}
    Under 
    Proposition~\ref{prop:consistencya}, let $\mathcal{G}^{\star}$ be the true AG. 
    Then, there are positive integers $N_{r}$ for each $r \in \mathbf{V}^{(2)}$ such that the expert-refined policy in Definition~\ref{def:experta} almost surely satisfies, after $N_{r}$ feedback instances for each variable pair $r$,
    \begin{equation*}
        p_{\top}^{\mathrm{ER}}(\mathcal{G}^{\star}) \coloneqq \sum_{\tau \colon s_{o} \rightsquigarrow \mathcal{G}^{\star}} p_{F}^{\mathrm{ER}}(\tau|s_{o}) > p_{\top}^{\mathrm{ER}}(\mathcal{G})
    \end{equation*}
    for any $\mathcal{G} \in \mathcal{X}$ 
    s.t. $\mathcal{G} \neq \mathcal{G}^{\star}$.
    Also, when $N_{r} \rightarrow \infty$ for $r \in \mathbf{V}^{(2)}$,
    $p_{\top}^{\mathrm{ER}}(\mathcal{G}^{\star}) \rightarrow 1$
    almost surely. \looseness=-1
\end{corollary}

In view of Corollary~\ref{col:consistencya}, we 
emphasize the separate roles of AGFN and the expert model in our EITL pipeline. 
On the one hand, AGFN provides the \emph{structural constraints} and a \emph{prior} for guided search on the space of AGs.
On the other hand, the expert provides \emph{local information} on the ancestral relationships between variable pairs.
Our work combines the strengths of both approaches to search for \emph{data-compatible} and \emph{expert-aligned} causal graphs.

Importantly, the feedback responses $f_{r}$ can only be collected by querying an expert (or many).
When interacting with the expert is expensive (e.g., due to the high cost of LLM API calls), this raises the question: which relation $r$ should we prioritize for building our dataset $\mathbf{f}_{r}$?
This question is addressed in the following section.

\subsection{Active knowledge elicitation}

Ideally, we would query the expert on the sequence of relations, $\{r_{i}\}_{i\ge1}$ in $\mathbf{V}^{(2)}$ that maximizes (in expectation) the convergence rate to the ground-truth AG in Corollary~\ref{col:consistencya}.
Nonetheless, this cannot be done due to the computational intractability of $p_{\top}^{\mathrm{ER}}(\mathcal{G}^{\star})$.
Instead, inspired by the active learning literature \cite{cohn1994active, bharti22human}, we propose probing the expert on the relationship that minimizes the expected entropy of our posterior in Lemma~\ref{lemma:posteriora}. \looseness=-1

\begin{definition}[Active knowledge elicitation]
    Let $\mathbf{f}_{r}^{N_{r}}$ be the $N_{r}$-sized set of feedback instances collected so far for relation $r \in \mathbf{V}^{(2)}$.
    The \emph{entropy} of $q_{r}(\cdot | \mathbf{f}_{r}^{N_{r}})$ is defined by $\mathbf{H}[q_{r}(\cdot|\mathbf{f}_{r}^{N_{r}})] = - \sum_{c \in \{\emptyset, \leftarrow, \rightarrow, \leftrightarrow\}} q_{r}(c | \mathbf{f}_{r}^{N_{r}}) \log q_{r}(c|\mathbf{f}_{r}^{N_{r}})$.
    We query the expert on the variable pair $r$ that minimizes the expected $\mathbf{H}[q_{r}(\cdot|\mathbf{f}_{r}^{N_{r}})]$ according to the generative model in \Cref{eq:bayesian}, that is, \looseness=-1 
    \begin{equation*}
        r^{\star} \coloneqq \argmin_{r \in \mathbf{V}^{(2)}} \mathbb{E}_{\pi, \omega} \mathbb{E}_{f_{r} \sim p(\cdot | \pi, \omega)} \left [ \mathbf{H}[q_{r}(\cdot |\mathbf{f}_{r}^{N_{r}} \cup \{f_{r}\})] \right],
    \end{equation*}
    in which the outer expectation is estimated with respect to the prior on $\pi$ and $\omega$ (either in \Cref{eq:bayesian} or in \Cref{eq:bayesianpriors}) and the inner expectation is computed with respect to to the Categorical distribution defining the expert's response $f_{r}$.
    \looseness=-1
\end{definition}

\begin{algorithm}[!t]
\caption{Active knowledge elicitation from the expert.}\label{alg:elicitations}
\begin{algorithmic}[1] 
\Require set of variable pairs $\mathbf{V}^{(2)} \coloneqq \binom{\mathbf{V}}{2}$
\Require $\mathbf{f}_{r}^{N_{r}} \in \{\emptyset, \leftarrow, \rightarrow, \leftrightarrow\}^{N_{r}}$ feedback instances received so far for $r \in \mathbf{V}^{(2)}$

\Procedure{GetExpectedEntropy}{$r, q_{r}, \mathbf{f}_{r}^{N_{r}}$}
    \State $(\omega_{1}, \pi_{1}), \dots, (\omega_{K}, \pi_{K}) \sim \mathrm{Prior}_{r}(\omega, \pi)$
    \State $E \gets 0$
    \For{$k \in \{1, \dots, K\}$}
        \For{$f_{r} \in \{\emptyset, \leftarrow, \rightarrow, \leftrightarrow\}$}
            \State $p_{k} \gets p(f_{r} | \pi_{k}, \omega_{k})$ \Comment{Compute the likelihood.}
            \State $E \gets E + p_{k} \cdot \mathbf{H}[q_{r}(\cdot | \mathbf{f}_{r}^{N_{r}} \cup \{f_{r}\})]$
        \EndFor
    \EndFor
    \State \Return $\nicefrac{E}{K}$ \Comment{Monte Carlo (exact calculation can be carried out for~\eqref{eq:bayesian}).}
\EndProcedure

\Procedure{SelectVariablePair}{$\{\mathbf{f}_{r}^{N} \colon r \in \mathbf{V}^{(2)}\}$} 
    \State $q_{r} \gets $ Compute posteriors from Lemma~\ref{lemma:posteriora} for $r \in \mathbf{V}^{(2)}$.
    \State $H \gets \{\}$
    \For{$r \in \mathbf{V}^{(2)}$}
        \State $H[r] \gets \textsc{GetExpectedEntropy}(r, q_{r}, \mathbf{f}_{r}^{N_{r}})$
    \EndFor
    \State $r^{\star} \gets \argmin_{r \in \mathbf{V}^{(2)}} H[r]$
    \State \Return $r^{\star}$
\EndProcedure


\Procedure{QueryExpert}{$V_{i}, V_{j}$}
    \Comment{Black-box function.}
    \State \Return $f_{\{V_{i}, V_{j}\}}$
\EndProcedure

\Procedure{UpdatePosterior}{$\{\mathbf{f}_{r}^{N_{r}} \colon r \in \mathbf{V}^{(2)}\}$}
    \State $r^{\star} \gets \textsc{SelectVariablePair}(\{\mathbf{f}_{r}^{N_{r}} \colon r \in \mathbf{V}^{(2)}\})$
    \State $f_{r^{\star}} \gets \textsc{QueryExpert}(r^{\star})$ \Comment{Recall $r^{\star} = \{V_{i}, V_{j}\}$ for certain $V_{i}, V_{j} \in \mathbf{V}$.}
    \State $N_{r^{\star}} \gets N_{r^{\star}} + 1$, $\mathbf{f}_{r^{\star}}^{N_{r^{\star}}} \gets \mathbf{f}_{r^{\star}}^{N_{r^{\star}}} \cup \{f_{r^{\star}}\}$
    \State \Return $\{\mathbf{f}_{r}^{N_{r}} \colon r \in \mathbf{V}^{(2)}\}$
\EndProcedure


\end{algorithmic}
\end{algorithm}

We found, during early experiments, that this strategy for actively interacting with the expert significantly accelerated the convergence of AGFN's mode to the AG when compared against a 
procedure that picks a relation $r$ uniformly at random. 
That said, we believe that
the design of 
enhanced elicitation strategies for EITL CD is an important direction for future research aiming to minimize the cost of expert interaction.
In \Cref{sec:experiments}, we provide an empirical assessment of AGFN using both simulated humans and LLMs as expert surrogates.
For reference, 
we also summarize 
our EITL framework in \Cref{alg:elicitations}. \looseness=-1


\section{Related Works}\label{sec:app:relat_work}

Before delving into our empirical analysis, we briefly overview the literature on CD under latent confounding and on expert-aided CD.
Interested readers may also consult \cite{spirtes2001causation, peters2017elements}. \looseness=-1

\subsection{CD under latent confounding} 
Following the seminal works by \citet{spirtes2001causation} and \citet{zhang2008fci} introducing the complete FCI, a variety of works have emerged. Among them are algorithms designed for sparse scenarios, including RFCI \citep{colombo2012rfci} and others \citep{silva2013mcmc, claassenMH13}. Notably,  \citet{silva2013mcmc}'s framework uses a Bayesian approach to CD of Gaussian causal diagrams based on sparse covariance matrices.
\citet{colombo2012rfci} introduced the conservative FCI to handle conflicts arising from statistical errors in scenarios with limited data, even though it yields less informative results.
Subsequent efforts to improve reliability led to the emergence of constraint-based CD algorithms based in Boolean satisfiability \citep{hyttinen2014constraint, magliacane2016ancestral}, although they are known to scale poorly on $|\mathbf{V}|$ \citep{lu2021improving}. In another paradigm, score-based search algorithms rank maximal AGs according to goodness-of-fit measures, commonly using BIC for linear Gaussian SCMs
\citep{triantafillou2016score, zhalama2017sat, DBLP:conf/uai/RantanenHJ21}. There are also hybrid approaches that combine constraint-based strategies to reduce the search space, such as GFCI~\citep{ogarrio2016gfci}, M3HC~\citep{tsirlis2018m3hc}, BCCD~\citep{classen2012bccd} and GSPo~\citep{bernstein2020ordering}.
Continuous optimization has recently emerged as a novel approach to score-based CD, as DCD \citep{bhattacharya2021differentiable} and N-ADMG \citep{ashman2023causal}. While N-ADMG focuses solely on finding bow-free causal diagrams, which are strictly less expressive than AGs, it offers some 
distributional estimation in the form of a variational posterior.
However, N-ADMG only samples a valid (bow-free) graph asymptotically \cite[Proposition 2]{ashman2023causal}, whereas AGFN exclusively generates AGs by design (Proposition~\ref{prop:agfnsa}).

\subsection{CD with expert knowledge}
Previous works on CD have explored various forms of background knowledge. This includes knowledge on edge existence/non-existence \citep{DBLP:conf/uai/Meek95a}, ancestral constraints \citep{ribeiro2024anchorfci, DBLP:conf/nips/ChenSCD16}, variable grouping \citep{DBLP:journals/ijar/ParviainenK17}, partial order \citep{DBLP:conf/aistats/Andrews20} and typing of variables \citep{DBLP:conf/clear2/BrouillardTLLD22}.  Incorporating expert knowledge is pivotal to reducing the search space and the size of the learned equivalence class. Nevertheless, due to significant challenges, up to date, there are only a few works trying to integrate background knowledge into CD within the context of latent confounding~\citep{DBLP:conf/aistats/Andrews20, DBLP:conf/nips/WangQZ22}. These works operate under the assumption of perfect expert feedback. In contrast, our contribution is the first to deal with more realistic situations where expert input might be inaccurate.
\looseness=-1

Recently, many works have been using the domain knowledge within LLMs to improve CD algorithms. Usually, the frameworks focus on using the LLM alongside a data-driven model.
To achieve this, most works elicit knowledge from the LLM in pairs of variables~\citep{llm_cd1,llm_cd4,eitl_uai2025}.
On top of that, recent works have also used LLMs for providing indirect information that improves the CD process.
For example,
i) \cite{junyiLLM2025} uses the LLM for intervention targeting, while ii) \citet{llm_cd5} solely focus on finding a causal order --- instead of the causal graph. Remarkably, none of the mentioned works consider the uncertainty inherent in answers given by the LLM, which we do.
\looseness=-1

\section{Experiments} \label{sec:experiments}

Our experimental analysis has two major objectives.
First, we demonstrate that AGFN approximates well a given distribution over AGs.
Second, we show that AGFN performs competitively with, and often drastically better than, strong baselines for causal discovery under latent confounding, namely FCI \citep{spirtes2001causation, zhang2008fci}, GFCI \citep{ogarrio2016gfci}, ACI \citep{magliacane2016ancestral}, DCD \citep{bhattacharya2021differentiable}, and N-ADMG \citep{ashman2023causal}.
We rigorously follow the protocols outlined in the publicly released code for the original works \cite{pcalga, pcalgaa, ogarrio2016gfci, ashman2023causal, magliacane2016ancestral, bhattacharya2021differentiable} when comparing them against AGFN. We refer readers to \Cref{sec:app:experimentsa} for experimental details.
Remarkably, we release the code for AGFN and our EITL framework in a \href{https://github.com/ML-FGV/agfn}{public GitHub repository} to support further research on expert-aided probabilistic CD. \looseness=-1

\subsection{Fitness to the target distribution} \label{sec:fitnessa}

\paragraph{Datasets}
To assess the distributional accuracy of AGFN, we randomly generate $5$-node AGs from a configuration model \cite{bollobas1980probabilistic, newman2010networks}.
For each AG, we generate a dataset of size 500 from a randomly parameterized linear Gaussian SCM that conforms to the AG's canonical causal diagram \cite{richardson2002ancestral}.
A canonical causal diagram is obtained from an AG by retaining all directed edges and replacing each bidirected edge $V_i \leftrightarrow V_j$ with a latent variable $U_{ij}$ that acts as a common cause of $V_i$ and $V_j$, representing latent confounding. For example, in Figure \ref{fig:ags_causaldiagrams}, the first ADMGs shown in the right panel correspond to the canonical causal diagrams of the respective AGs.
Based on these datasets, we compute the reward function (or target distribution) associated to each graph $\mathcal{G}$ as the negative exponential of the modified BIC score \cite{foygel2010extendedbayesianinformationcriteria}.
That is, denoting by $\mathcal{D}$ our dataset,
\begin{equation} \label{eq:bicsa}
    T \cdot \log R(\mathcal{G}) = -\mathrm{BIC}(\mathcal{G}, \mathcal{D}) = \textcolor{teal}{2 \ell( \hat{\boldsymbol{\theta}}(\mathcal{G},
    \mathcal{D}))} \textcolor{orange}{- |\mathcal{G}|_{E} \log |\mathcal{D}| - 2 |\mathcal{G}|_{E} \log |\mathcal{G}|_{N}},
\end{equation}
in which $\hat{\boldsymbol{\theta}}(\mathcal{D}, \mathcal{G})$ denotes the maximum likelihood parameter for the family of SCMs abiding by $\mathcal{G}$, which can be efficiently computed via the iterative residual least-squares algorithm (IRLS)\footnote{We provide a \texttt{Python}-based GPU implementation for IRLS in the released code.} \cite{drton2012iterativeconditionalfittinggaussian}, $|\mathcal{G}|_{N}$ and $|\mathcal{G}|_{E}$ represent the number of nodes and edges of $\mathcal{G}$, respectively, $|\mathcal{D}|$ the number of samples in our dataset ($500$, in this case), and $T > 0$ is a temperature parameter.
Intuitively, $R(\cdot)$ favors data-compatible AGs (\textcolor{teal}{teal}) while penalizing complex (dense) structures (\textcolor{orange}{orange}).
Importantly, the maximizer of $R(\cdot)$ is, under standard Gaussianity assumptions, a consistent estimator of the true 
structure $\mathcal{G}^{\star}$ \cite[Main Theorem]{foygel2010extendedbayesianinformationcriteria}. \looseness=-1

\begin{remark}[On the choice of a target distribution]
    From a broader viewpoint, \Cref{eq:bicsa} provides a blueprint for designing AGFN's target distribution: $R(\cdot)$ should assign high probability to AGs that fit well to the observed data and, ideally, enjoy certain asymptotic properties (such as consistency).
    As such, BIC could be replaced by alternative scoring functions from the vast literature on
    score-based model selection \cite{robert2007bayesian} as well as from recent works on scoring AGs for causal discovery \cite{andrews19alearning, lagrange2025efficient, ramsey2025scalablecausaldiscoveryrecursive, ribeiro2025dcfci}).
    In doing so, our objective is for the learned policy function $p_{F}$ to be biased toward data-compatible AGs, while retaining support for structures that are less consistent with the possibly unfaithful data.
    \looseness=-1
\end{remark}

Additionally, we train an AGFN to accurately sample \emph{sparse} 25-node AGs using data generated from SCMs that conform to the canonical causal diagram associated with each AG.
Notably, this scale of 25-node AGs substantially exceeds the problem sizes considered in prior work on amortized inference-based structure learning \cite{deleu2022bayesian, deleu2023joint}.
Furthermore, using datasets from the DREAM3 challenge \cite{madar2010dream3}, we demonstrate that AGFNs can efficiently identify high-scoring AGs in larger settings even when the target distribution is only imperfectly approximated.
In this way, we also contribute to the mounting evidence on the effectiveness of amortized samplers for combinatorial optimization problems \cite{zhang2023robust, Zhang2023}.
\looseness=-1

\begin{figure*}[t]
    \centering

    \begin{minipage}[t]{0.48\textwidth}
        \hspace{-24pt}
        \includegraphics[width=\linewidth]{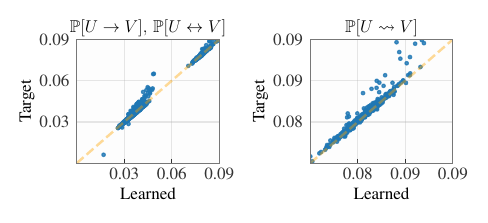}
        \caption{AGFN accurately learns a distribution over sparse (degree-bounded) 25-node AGs.}
        \label{fig:agfnsdistlarges}
    \end{minipage}%
    \hspace{6pt}
    \begin{minipage}[t]{0.42\textwidth}
        \vspace{-102pt}
        \hspace{-24pt}
        \begin{tabular}{c|c|c}
             & \textrm{Yeast} & \textrm{E. coli} \\
             \hline
            N-ADMG & $21762.50 { \scriptstyle \textcolor{gray}{ \pm 448.51 } }$ & $22217.61 { \scriptstyle \textcolor{gray}{ \pm 228.28 } }$ \\
            DCD & $20941.02 { \scriptstyle \textcolor{gray}{ \pm 127.62 } }$ & $21994.54 { \scriptstyle \textcolor{gray}{ \pm 138.08 } }$ \\
            GFCI & $21000.43 { \scriptstyle \textcolor{gray}{ \pm 183.23 } }$ & $21920.66 { \scriptstyle \textcolor{gray}{ \pm 135.51 } }$ \\
            FCI & $21250.00 { \scriptstyle \textcolor{gray}{ \pm 438.83 } }$ & $22102.09 { \scriptstyle \textcolor{gray}{ \pm 88.98 } }$ \\
            ACI & $21215.77 { \scriptstyle \textcolor{gray}{ \pm 246.04 } }$ & $22071.27 { \scriptstyle \textcolor{gray}{ \pm 188.83 } }$ \\
            AGFN & $\mathbf{20918.87} { \scriptstyle \textcolor{gray} { \pm 71.89 } }$ & $\mathbf{21852.23} { \scriptstyle \textcolor{gray} { \pm 45.29 } }$ \\
        \end{tabular}
        \captionof{table}{BIC for GRN datasets for both AGFN (\textbf{bold}) and baselines.
        Lower is better. \looseness=-1}e
        \label{tab:bics}
    \end{minipage}

    \label{fig:agfn_horizontal}
\end{figure*}

\paragraph{Setup}
For each dataset, we parameterize $p_{F}$ with an MLP having two 256-dimensional hidden layers and, as explained earlier in \Cref{sec:background}, let $p_{B}$ be an uniform policy.
Then, we train AGFN by minimizing the TB objective in Proposition~\ref{prop:tba} for $3 \cdot 10^{3}$ steps with the Adam \cite{kingma2017adammethodstochasticoptimization} optimizer and, for each step, we compute the loss function based on a batch of $64$ trajectories. \looseness=-1
When learning a distribution over sparse 25-node AGs, we constraint both the in- and out-degrees of the learned graphs to a maximum of 2.
As in \cite{deleu2022bayesian, deleu2023joint}, since directly sampling from $R$ is generally intractable due to the enormous size of the space of AGs (recall \Cref{fig:nodesa}), we assess AGFN's distributional accuracy by comparing its learned distribution against the target in \Cref{eq:bicsa} on the space of potential \emph{ancestral relationships}, i.e.,  \looseness=-1
\begin{equation} \label{eq:edgesa}
    \mathbb{P}[U r V] = \frac{1}{K} \sum_{1 \le k \le K} [ (U r V) \in \mathcal{G}_{k}],
\end{equation}
in which $r \in \{\leftarrow, \rightarrow, \leftrightarrow\}$ and $\{\mathcal{G}_{k}\}_{k=1}^{K}$ is a $K$-sized sample from $p_{F}$.
We approximate the ground-truth value of these probabilities on $\{\mathcal{G}_{k}\}_{k=1}^{K}$ via importance sampling \cite{kloek1978bayesian}.
\looseness=-1

\begin{figure*}
    \centering
    \includegraphics[width=.9\linewidth]{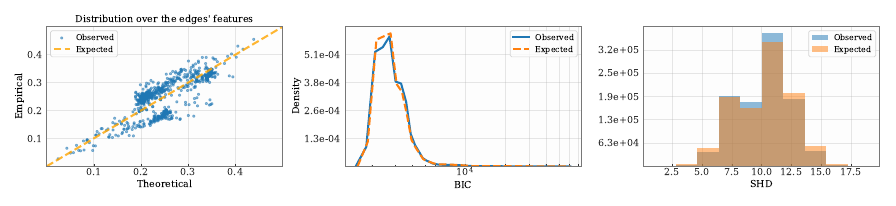}
    \caption{\textbf{AGFN accurately learns a distribution over AGs.} 
    AGFN's empirical distribution (\textcolor{blue}{blue}) over relationship types (left; see \Cref{eq:edgesa}), BIC (middle), and structural Hamming distance (SHD; right) approximately match their expected value based on the target distribution (\textcolor{orange}{orange}) defined in \Cref{eq:bicsa}.
    }
    \label{fig:agfnsdists}
\end{figure*}

\paragraph{Results}
\Cref{fig:agfnsdists} shows that AGFNs can accurately sample from a distribution over AGs.
Similarly, \Cref{fig:agfnsdistlarges} demonstrates our model's capability in sampling from a distribution over sparse AGs with 25 nodes.
There, we also notice that $\mathbb{P}[U \rightsquigarrow V]$, the probability that there is a directed path from variables $U$ to $V$ on the sampled AG, closely matches its expected value from \Cref{eq:bicsa}.
In conclusion, \Cref{tab:bics} highlights that AGFN is capable of finding higher-scoring AGs than strong baselines (described in \Cref{sec:app:relat_work}) on the 10-node gene regulatory networks (GRNs) datasets, each with 200 samples, from the DREAM3 challenge \cite{madar2010dream3}. \looseness=-1

\subsection{Expert-in-the-loop Causal Discovery}

\begin{figure*}
    \centering
    \includegraphics[width=0.75\linewidth]{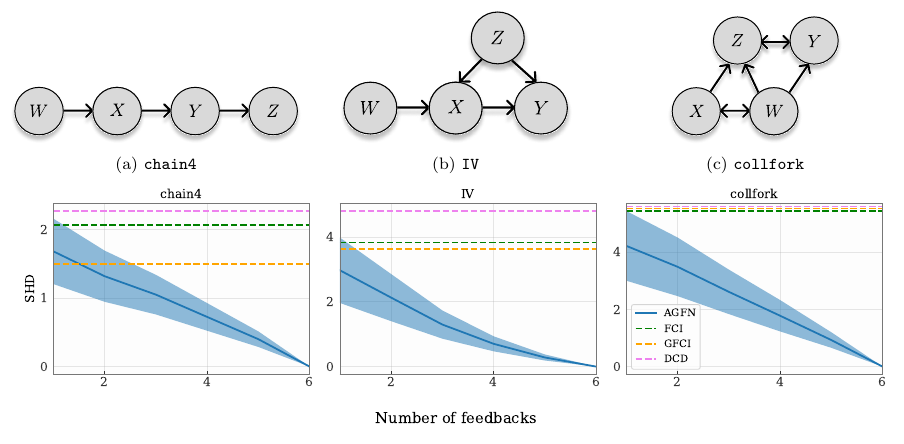}
    \caption{(Top, a-c) AGs with standard structural motifs (chain, collider, latent confounder) which we use to evaluate our EITL framework. (Bottom) AGFN finds a more accurate AG than strong baselines. Results for expert-refined AGFN in terms of the SHD as a function of the number of responses from an uncertain expert (according to \Cref{eq:bayesian}). Error bars represent standard deviations across independent expert simulations.
    Dashed horizontal lines indicate the SHD of the considered baselines.}
    \label{fig:expertsa}
\end{figure*}

\paragraph{Datasets}
We evaluate our EITL framework on datasets abiding by four distinct AGs.
We first consider the three AGs shown in \Cref{fig:expertsa}. While relatively simple, these AGs capture progressively more complex causal structures and are used to evaluate the ability to recover key structural components: \texttt{chain4} contains only directed edges, \texttt{iv} includes collider nodes, and \texttt{collfork} features bidirected edges representing associations induced solely by latent confounding.
As in the previous section, we generate $500$-sized datasets from a randomly parameterized linear Gaussian SCM abiding by the canonical causal diagram of each AG.
We also present results for the Sachs dataset \cite{sachs} using an LLM as the expert, where we marginalize a subset of the variables to introduce latent confounding (see \Cref{sec:app:experimentsa} for more details). \looseness=-1

\paragraph{Setup}
We follow the setup in \Cref{sec:fitnessa} for training our AGFN.
In particular, we also use the modified BIC in \Cref{eq:bicsa} as the target distribution.
For the synthetic AGs in \Cref{fig:expertsa}(a-c), we evaluate our EITL framework with a simulate human abiding by \Cref{eq:bayesian} with $\pi_{r} = 0.8$ for each variable pair $r$.
On the other hand, we use GPT-4o \cite{openai2024gpt4technicalreport} as an expert for the Sachs dataset; we refer the reader to the supplement for details on prompting and implementation.
As in \Cref{def:experta}, we choose the best AG based on the log-pooling between $R(\cdot)$ and the expert-induced 
distribution over AGs, computed as the edge-wise product of the corresponding posterior in \Cref{eq:bayesian}.
We use \Cref{eq:bayesian} for all experiments.
\looseness=-1

\begin{figure*}[!t]
    \centering
    \includegraphics[width=0.75\textwidth]{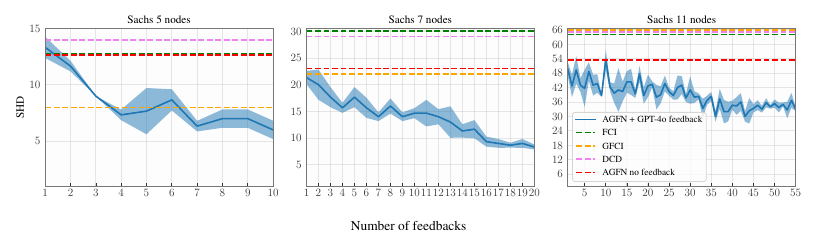}
    \caption{\textbf{LLM-aided AGFN outperforms CD baselines} for the Sachs dataset. The plots show the average and standard deviation of the SHD computed across independent expert interactions.
    We also evaluate the considered algorithms to an marginalized version of the Sachs dataset, in which we remove either 6 (left) or 4 (middle) variables.
    In doing so, we reduce the search space, but introduce latent confounding. \looseness=-1
    }
    \vspace{-1em}
    \label{fig:graph_sachs}
\end{figure*}

\paragraph{Results}
We compare an expert-refined AGFN against FCI \cite{zhang2008fci}, GFCI \cite{ogarrio2016gfci}, and DCD \cite{bhattacharya2021differentiable}, omitting ACI results \cite{magliacane2016ancestral} in \Cref{fig:graph_sachs,fig:expertsa} as they are significantly larger than the ones we present.
Notably, our model finds a more accurate causal structure than any of the considered baselines for both synthetic and realistic datasets---even after receiving only a relatively small number of expert responses.
All in all, these observations underline the potential of AGFN as a probabilistic EITL CD for possibly latently confounded datasets.
\looseness=-1

\section{Conclusions} \label{sec:conclusionsa}

We introduced AGFN as the first amortized sampler for the space of AGs.
Drawing on this, we developed the first probabilistic EITL CD algorithm under latent confounding.
Importantly, we showed that AGFN is amenable to the incorporation of ex-ante structural background knowledge, such as sparsity, partitionability, and unconfoundedness---which we have only partially explored in the experiments---, and to ex-post information from a noisy expert.
We also proved the consistency of our Bayesian model when the expert's feedback is better-than-random, which holds even when the likelihood function is mildly misspecified.
In closing, we provided a standard empirical analysis that validates our theory and attests the effectiveness of AGFN for EITL CD in both synthetic and realistic latently confounded datasets.
\looseness=-1

That said, AGFN is highly modular, and many of its components warrant further exploration in future research.
For instance, is our entropy-minimizing elicitation strategy the optimal approach for interacting with the expert in some sense?
Is log-pooling the ideal 
mechanism for combining the expert-induced posterior with AGFN's learned policy function?
Which neural network architecture minimizes the sample complexity for learning an accurate policy function for AGFN?
Our Bayesian hierarchical scheme assumed that the observed feedback responses are conditionally independent.
How to handle serially dependent responses, i.e., a time series?
When evaluating search methods that leverage LLM knowledge on public benchmarks, we also face the risk of leakage of the task's response to the LLM during its pretraining.
As such, how can we fairly evaluate CD algorithms that rely on LLMs?
Notably, our current approach requires retraining for each dataset.
Can we construct a general-purpose AGFN that amortizes not only over the state graph (via the policy function), but also over datasets?
This is also connected to an emerging literature on the design of foundation models for probabilistic inference \cite{chang2025amortizedprobabilisticconditioningoptimization, hassan2025efficient}. \looseness=-1

While challenges in CD persist, AGFN's integration of noisy human expertise into the search process is a key step towards human-aligned and data-efficient causal discovery. \looseness=-1

\section*{Data statement}

Our experiments used simulated (including \texttt{chain4}, \texttt{IV}, and \texttt{collfork}) and realistic (DREAM3 \cite{madar2010dream3} and Sachs \cite{sachs}) datasets.
We release the computer code for data generation in our public GitHub repository.
Also, DREAM3 and Sachs are publicly available online.

\section*{Author contributions: CRediT}
\textbf{Tiago da Silva}: Conceptualization, Software, Formal analysis, Writing - Original Draft, Writing - Review \& Editing, Visualization; \textbf{Bruna Bazaluk}: Conceptualization, Software, Writing - Original Draft, Writing - Review \& Editing, Visualization; \textbf{Eliezer de Souza da Silva}: Conceptualization, Writing - Original Draft, Writing - Review \& Editing, Visualization; \textbf{António Góis}: Conceptualization, Writing - Original Draft, Writing - Review \& Editing, Visualization; \textbf{Salem Lahlou}: Writing - Review \& Editing, Supervision; \textbf{Dominik Heider}: Writing - Review \& Editing, Resources, Funding acquisition; \textbf{Samuel Kaski}: Writing - Review \& Editing, Resources; \textbf{Diego Mesquita}: Conceptualization, Writing - Original Draft, Writing - Review \& Editing, Visualization, Supervision, Resources; \textbf{Adèle Helena Ribeiro}: Conceptualization, Software, Writing - Original Draft, Writing - Review \& Editing, Visualization, Supervision. \looseness=-1

\section*{Funding Sources}
Tiago da Silva, Eliezer Silva, and Diego Mesquita acknowledge the support of Fundação Carlos Chagas Filho de Amparo à Pesquisa do Estado do Rio de Janeiro FAPERJ (SEI-260003/000709/2023), São Paulo Research Foundation (FAPESP, grant 2023/00815-6), Conselho Nacional de Desenvolvimento Científico e Tecnológico (CNPq, grant 404336/2023-0).
Bruna Bazaluk acknowledges the support of Coordenação de Aperfeiçoamento de Pessoal de Nível Superior – Brasil (CAPES) – Código de Financiamento 001.
Adèle H. Ribeiro and Dominik Heider were supported by the LOEWE program of the State of Hesse (Germany) in the Diffusible Signals research cluster and by the German Federal Ministry of Research, Technology and Space of Germany (BMFTR) [grant number 031L0267A] (Deep Insight). Adèle H. Ribeiro was further supported by BMFTR [grant number 01ZU2503] (CausalAI4Health).
António Góis was supported by Samsung Electronics Co., Ltd.
Samuel Kaski was supported by the Academy of Finland (Flagship programme: Finnish Center for Artificial Intelligence FCAI), EU Horizon 2020 (European Network of AI Excellence Centres ELISE, grant agreement 951847), UKRI Turing AI World-Leading Researcher Fellowship (EP/W002973/1).
We also acknowledge the computational resources provided i) by the Aalto Science-IT Project from Computer Science IT, and ii) FGV TIC.

\bibliography{biblio}

@inproceedings{
zhang2023robust,
title={Robust Scheduling with {GF}lowNets},
author={David W Zhang and Corrado Rainone and Markus Peschl and Roberto Bondesan},
booktitle={International Conference on Learning Representations (ICLR)},
year={2023},
}

@book{spirtes2001causation,
    title = {{Causation, Prediction, and Search}},
    year = {2001},
    author = {Spirtes, Peter and Glymour, Clark N and Scheines, Richard},
    edition = {2nd},
    publisher = {MIT Press},
}

@inproceedings{ashman2023causal,
title={Causal Reasoning in the Presence of Latent Confounders via Neural {ADMG} Learning},
author={Matthew Ashman and Chao Ma and Agrin Hilmkil and Joel Jennings and Cheng Zhang},
booktitle={International Conference on Learning Representations (ICLR)},
year={2023},
}

@inproceedings{DBLP:conf/uai/Meek95a,
  author       = {Christopher Meek},
  title        = {Strong completeness and faithfulness in Bayesian networks},
  booktitle    = {Uncertainty in Artificial Intelligence (UAI)},
  year         = {1995}
}

@article{colombo2014order,
  title={Order-independent constraint-based causal structure learning.},
  author={Colombo, Diego and Maathuis, Marloes H and others},
  journal={J. Mach. Learn. Res.},
  volume={15},
  number={1},
  pages={3741--3782},
  year={2014}
}

@article{zhang2016faces,
author = {Zhang, Jiji and Spirtes, Peter},
doi = {10.1007/s11229-015-0673-9},
issn = {1573-0964},
journal = {Synthese},
number = {4},
pages = {1011--1027},
title = {{The three faces of faithfulness}},
volume = {193},
year = {2016}
}

@article{zhalama2017heuristic,
author = {Zhalama and Zhang, Jiji and Mayer, Wolfgang},
doi = {10.1007/s41060-016-0033-y},
issn = {2364-4168},
journal = {International Journal of Data Science and Analytics},
number = {2},
pages = {93--104},
title = {{Weakening faithfulness: some heuristic causal discovery algorithms}},
volume = {3},
year = {2017}
}

@article{Zhang2008,
    author= {'J. Zhang and P. Spirtes{\color{white}'\!}},
    title={Detection of Unfaithfulness and Robust Causal Inference},
    journal={Minds and Machines},
    year={2008},
    month={Jun},
    volume={18},
    number={2},
    pages={239--271},
}

@article{zhang2008fci,
  author       = {Jiji Zhang},
  title        = {On the completeness of orientation rules for causal discovery in the presence of latent confounders and selection bias},
  journal      = {Artificial Intelligence},
  year         = {2008}
}

@inproceedings{DBLP:conf/iclr/LiLSH23,
  author       = {Yinchuan Li and
                  Shuang Luo and
                  Yunfeng Shao and
                  Jianye Hao},
  title        = {GFlowNets with Human Feedback},
  booktitle    = {Tiny Papers @ {(ICLR)}},
  publisher    = {OpenReview.net},
  year         = {2023}
}

@article{DBLP:journals/ai/WangQZ23,
  author       = {Tian{-}Zuo Wang and
                  Tian Qin and
                  Zhi{-}Hua Zhou},
  title        = {Sound and complete causal identification with latent variables given
                  local background knowledge},
  journal      = {Artif. Intell.},
  volume       = {322},
  pages        = {103964},
  year         = {2023}
}

@inproceedings{DBLP:conf/nips/WangQZ22,
  author       = {Tian{-}Zuo Wang and
                  Tian Qin and
                  Zhi{-}Hua Zhou},
  title        = {Sound and Complete Causal Identification with Latent Variables Given
                  Local Background Knowledge},
  booktitle    = {Advances in Neural Information Processing Systems (NeurIPS)},
  year         = {2022}
}

@inproceedings{DBLP:conf/aistats/Andrews20,
  author       = {Bryan Andrews},
  title        = {On the Completeness of Causal Discovery in the Presence of Latent
                  Confounding with Tiered Background Knowledge},
  booktitle    = {Artificial Intelligence and Statistics (AISTATS)},
  year         = {2020}
}

@inproceedings{DBLP:conf/clear2/BrouillardTLLD22,
  author       = {Philippe Brouillard and
                  Perouz Taslakian and
                  Alexandre Lacoste and
                  S{\'{e}}bastien Lachapelle and
                  Alexandre Drouin},
  title        = {Typing assumptions improve identification in causal discovery},
  booktitle    = {Causal Learning and Reasoning (CLeaR)},
  year         = {2022}
}

@inproceedings{DBLP:conf/nips/ChenSCD16,
  author       = {Eunice Yuh{-}Jie Chen and
                  Yujia Shen and
                  Arthur Choi and
                  Adnan Darwiche},
  title        = {Learning Bayesian networks with ancestral constraints},
  booktitle    = {Advances in Neural Information Processing Systems (NeurIPS)},
  year         = {2016}
}

@article{DBLP:journals/ijar/ParviainenK17,
  author       = {Pekka Parviainen and
                  Samuel Kaski},
  title        = {Learning structures of Bayesian networks for variable groups},
  journal      = {Int. J. Approx. Reason.},
  volume       = {88},
  pages        = {110--127},
  year         = {2017}
}

@article{DBLP:journals/ijar/HauserB14,
  author       = {Alain Hauser and
                  Peter B{\"{u}}hlmann},
  title        = {Two optimal strategies for active learning of causal models from interventional
                  data},
  journal      = {Int. J. Approx. Reason.},
  volume       = {55},
  number       = {4},
  pages        = {926--939},
  year         = {2014}
}

@inproceedings{claassen2022greedy,
  title={Greedy equivalence search in the presence of latent confounders},
  author={Claassen, Tom and Bucur, Ioan G},
  booktitle={Uncertainty in Artificial Intelligence},
  pages={443--452},
  year={2022},
  organization={PMLR}
}

@inproceedings{hyttinen2014constraint,
  title={Constraint-based Causal Discovery: Conflict Resolution with Answer Set Programming.},
  author={Hyttinen, Antti and Eberhardt, Frederick and J{\"a}rvisalo, Matti},
  booktitle={Uncertainty in Artificial Intelligence (UAI)},
  year={2014}
}

@inproceedings{lu2021improving,
  title={Improving causal discovery by optimal bayesian network learning},
  author={Lu, Ni Y and Zhang, Kun and Yuan, Changhe},
  booktitle={AAAI Conference on Artificial Intelligence (AAAI)},
  year={2021}
}

@inproceedings{DBLP:conf/uai/RantanenHJ21,
  author       = {Kari Rantanen and
                  Antti Hyttinen and
                  Matti J{\"{a}}rvisalo},
  title        = {Maximal ancestral graph structure learning via exact search},
  booktitle    = {Artificial Intelligence and Statistics (UAI)},
  year         = {2021}
}

@article{zhang2008causal,
author = {Zhang, Jiji},
title = {Causal Reasoning with Ancestral Graphs},
year = {2008},
issue_date = {6/1/2008},
publisher = {JMLR.org},
volume = {9},
issn = {1532-4435},
journal = {J. Mach. Learn. Res.},
month = jun,
pages = {1437–1474},
numpages = {38}
}

@Book{pearl:2k,
  author =   "Pearl, Judea",
  title =    "Causality: Models, Reasoning, and Inference",
  year =     2000,
  publisher =    "Cambridge University Press",
  address =      "New York",
  note =     "2nd edition, 2009"
}

@article{ng2021reliable,
  title={Reliable causal discovery with improved exact search and weaker assumptions},
  author={Ng, Ignavier and Zheng, Yujia and Zhang, Jiji and Zhang, Kun},
  journal={Advances in Neural Information Processing Systems (NeurIPS)},
  year={2021}
}

@inproceedings{claassenMH13,
  author       = {Tom Claassen and
                  Joris M. Mooij and
                  Tom Heskes},
  title        = {Learning Sparse Causal Models is not NP-hard},
  booktitle    = {Uncertainty in Artificial
                  Intelligence (UAI)},
  year         = {2013}
}

@article{colombo2012rfci,
author = {Colombo, Diego and Maathuis, Marloes H and Kalisch, Markus and Richardson, Thomas S},
journal = {Annals of Statistics},
title = {{Learning high-dimensional directed acyclic graphs with latent and selection variables}},
year = {2012}
}

@inproceedings{zhalama2017sat,
  author       = {Zhalama and
                  Jiji Zhang and
                  Frederick Eberhardt and
                  Wolfgang Mayer},
  title        = {SAT-Based Causal Discovery under Weaker Assumptions},
  booktitle    = {Artificial Intelligence and Statistics {(UAI)}},
  publisher    = {{AUAI} Press},
  year         = {2017}
}

@inproceedings{DBLP:conf/uai/Zhang07,
  author       = {Jiji Zhang},
  title        = {A Characterization of Markov Equivalence Classes for Directed Acyclic
                  Graphs with Latent Variables},
  booktitle    = {Artificial Intelligence and Statistics {(UAI)}},
  pages        = {450--457},
  publisher    = {{AUAI} Press},
  year         = {2007}
}

@InProceedings{andrews19alearning,
  title = 	 {Learning High-dimensional Directed Acyclic Graphs with Mixed Data-types},
  author =       {Andrews, Bryan and Ramsey, Joseph and Cooper, Gregory F.},
  booktitle = 	 {Proceedings of Machine Learning Research},
  pages = 	 {4--21},
  year = 	 {2019},
  editor = 	 {},
  volume = 	 {104},
  series = 	 {Proceedings of Machine Learning Research},
  month = 	 {05 Aug},
  publisher =    {PMLR},
  url = 	 {https://proceedings.mlr.press/v104/andrews19a.html}
}

@inproceedings{deleu2022bayesian,
  title={Bayesian Structure Learning with Generative Flow Networks},
  author={Deleu, Tristan and G{\'o}is, Ant{\'o}nio and Emezue, Chris Chinenye and Rankawat, Mansi and Lacoste-Julien, Simon and Bauer, Stefan and Bengio, Yoshua},
  booktitle={Uncertainty in Artificial Intelligence (UAI)},
  year={2022}
}

@inproceedings{classen2012bccd,
author = {Claassen, Tom and Heskes, Tom},
title = {A Bayesian Approach to Constraint Based Causal Inference},
year = {2012},
booktitle = {Uncertainty in Artificial Intelligence (UAI)},
}

@incollection{silva2013mcmc,
  author       = {Ricardo Silva},
  title        = {A {MCMC} Approach for Learning the Structure of Gaussian Acyclic Directed
                  Mixed Graphs},
  booktitle    = {Statistical Models for Data Analysis},
  series       = {Studies in Classification, Data Analysis, and Knowledge Organization},
  pages        = {343--351},
  publisher    = {Springer},
  year         = {2013}
}

@article{bengio2021gflownet,
  title={{Flow Network based Generative Models for Non-Iterative Diverse Candidate Generation}},
  author={Bengio, Emmanuel and Jain, Moksh and Korablyov, Maksym and Precup, Doina and Bengio, Yoshua},
  journal={Advances in Neural Information Processing Systems (NeurIPS)},
  year={2021}
}

@article{bengio2021gflownetfoundations,
  title={{GFlowNet Foundations}},
  author={Bengio, Yoshua and Deleu, Tristan and Hu, Edward J and Lahlou, Salem and Tiwari, Mo and Bengio, Emmanuel},
  journal={arXiv preprint},
  year={2021}
}

@article{tsirlis2018m3hc,
author = {Tsirlis, Konstantinos and Lagani, Vincenzo and Triantafillou, Sofia and Tsamardinos, Ioannis},
issn = {0888-613X},
journal = {International Journal of Approximate Reasoning},
pages = {74--85},
title = {{On scoring Maximal Ancestral Graphs with the Max–Min Hill Climbing algorithm}},
volume = {102},
year = {2018}
}

@inproceedings{bernstein2020ordering,
  title={Ordering-based causal structure learning in the presence of latent variables},
  author={Bernstein, Daniel and Saeed, Basil and Squires, Chandler and Uhler, Caroline},
  booktitle={Artificial Intelligence and Statistics (AISTATS)},
  year={2020}
}

@article{richardson2002ancestral,
  title={Ancestral graph Markov models},
  author={Richardson, Thomas and Spirtes, Peter},
  journal={Annals of Statistics},
  year={2002},
  publisher={Institute of Mathematical Statistics}
}

@inproceedings{bhattacharya2021differentiable,
  title={Differentiable causal discovery under unmeasured confounding},
  author={Bhattacharya, Rohit and Nagarajan, Tushar and Malinsky, Daniel and Shpitser, Ilya},
  booktitle={Artificial Intelligence and Statistics (AISTATS)},
  year={2021},
}

@article{magliacane2016ancestral,
  title={Ancestral causal inference},
  author={Magliacane, Sara and Claassen, Tom and Mooij, Joris M},
  journal={Advances in Neural Information Processing Systems (NeurIPS)},
  year={2016}
}

@InProceedings{ogarrio2016gfci,
  title = 	 {A Hybrid Causal Search Algorithm for Latent Variable Models},
  author = 	 {Ogarrio, Juan Miguel and Spirtes, Peter and Ramsey, Joe},
  booktitle = 	 {Probabilistic Graphical Models (PGM)},
  year = 	 {2016},
}

@inproceedings{anand2023causal,
  title={Causal effect identification in cluster dags},
  author={Anand, Tara V and Ribeiro, Adele H and Tian, Jin and Bareinboim, Elias},
  booktitle={Proceedings of the AAAI Conference on Artificial Intelligence},
  volume={37},
  number={10},
  pages={12172--12179},
  year={2023}
}

@inproceedings{anand2025cd,
  title={Causal Discovery over Clusters of Variables in Markovian Systems},
  author={Anand, Tara Vafai and Ribeiro, Ad{\`e}le H and Tian, Jin and Hripcsak, George and Bareinboim, Elias},
  booktitle={The Thirty-ninth Annual Conference on Neural Information Processing Systems},
  year={2025}
}

@incollection{ribeiro2016causal,
  title={Causal inference and structure learning of genotype--phenotype networks using genetic variation},
  author={Ribeiro, Ad{\`e}le H and Soler, J{\'u}lia MP and Neto, Elias Chaibub and Fujita, Andr{\'e}},
  booktitle={Big data analytics in genomics},
  pages={89--143},
  year={2016},
  publisher={Springer}
}

@article{ribeiro2025dcfci,
  title={dcFCI: Robust Causal Discovery Under Latent Confounding, Unfaithfulness, and Mixed Data},
  author={Ribeiro, Ad{\`e}le H and Heider, Dominik},
  journal={arXiv preprint arXiv:2505.06542},
  year={2025}
}

@inproceedings{triantafillou2016score,
  title={Score-based vs Constraint-based Causal Learning in the Presence of Confounders},
  author={Triantafillou, Sofia and Tsamardinos, Ioannis},
  booktitle={Causation: Foundation to Application  Workshop (CFA)},
  pages={59--67},
  year={2016}
}

@article{knox2024drugbank,
  title={Drugbank 6.0: the drugbank knowledgebase for 2024},
  author={Knox, Craig and Wilson, Mike and Klinger, Christen M and Franklin, Mark and Oler, Eponine and Wilson, Alex and Pon, Allison and Cox, Jordan and Chin, Na Eun and Strawbridge, Seth A and others},
  journal={Nucleic Acids Research},
  volume={52},
  number={D1},
  pages={D1265--D1275},
  year={2024},
  publisher={Oxford University Press}
}

@article{wheeler2007database,
  title={Database resources of the national center for biotechnology information},
  author={Wheeler, David L and Barrett, Tanya and Benson, Dennis A and Bryant, Stephen H and Canese, Kathi and Chetvernin, Vyacheslav and Church, Deanna M and DiCuccio, Michael and Edgar, Ron and Federhen, Scott and others},
  journal={Nucleic acids research},
  volume={36},
  number={suppl\_1},
  pages={D13--D21},
  year={2007},
  publisher={Oxford University Press}
}

@article{thomas2019gene,
  title={Gene Ontology Causal Activity Modeling (GO-CAM) moves beyond GO annotations to structured descriptions of biological functions and systems},
  author={Thomas, Paul D and Hill, David P and Mi, Huaiyu and Osumi-Sutherland, David and Van Auken, Kimberly and Carbon, Seth and Balhoff, James P and Albou, Laurent-Philippe and Good, Benjamin and Gaudet, Pascale and others},
  journal={Nature genetics},
  volume={51},
  number={10},
  pages={1429--1433},
  year={2019},
  publisher={Nature Publishing Group US New York}
}

@InProceedings{bharti22human,
  title = 	 {Approximate {B}ayesian Computation with Domain Expert in the Loop},
  author =       {Bharti, Ayush and Filstroff, Louis and Kaski, Samuel},
  booktitle = 	 {International Conference on Machine Learning (ICML)},
  year = 	 {2022},
}

@article{malkin2022trajectory,
  title={Trajectory balance: Improved credit assignment in gflownets},
  author={Malkin, Nikolay and Jain, Moksh and Bengio, Emmanuel and Sun, Chen and Bengio, Yoshua},
  journal={Advances in Neural Information Processing Systems (NeurIPS)},
  volume={35},
  pages={5955--5967},
  year={2022}
}

@inproceedings{lahlou2023continuous,
  title={A theory of continuous generative flow networks},
  author={Lahlou, Salem and Deleu, Tristan and Lemos, Pablo and Zhang, Dinghuai and Volokhova, Alexandra and Hern{\'a}ndez-Garc{\i}a, Alex and Ezzine, L{\'e}na N{\'e}hale and Bengio, Yoshua and Malkin, Nikolay},
  booktitle={International Conference on Machine Learning},
  pages={18269--18300},
  year={2023},
  organization={PMLR}
}

@article{deleu2023joint,
  title={Joint Bayesian Inference of Graphical Structure and Parameters with a Single Generative Flow Network},
  author={Deleu, Tristan and Nishikawa-Toomey, Mizu and Subramanian, Jithendaraa and Malkin, Nikolay and Charlin, Laurent and Bengio, Yoshua},
  journal={arXiv preprint arXiv:2305.19366},
  year={2023}
}

@article{ryan15bayesian,
  year = {2015},
  volume = {84},
  number = {1},
  pages = {128--154},
  author = {Elizabeth G. Ryan and Christopher C. Drovandi and James M. McGree and Anthony N. Pettitt},
  title = {A Review of Modern Computational Algorithms for Bayesian Optimal Design},
  journal = {International Statistical Review}
}

@article{torch,
  author       = {Adam Paszke and
                  Sam Gross and
                  Francisco Massa and
                  Adam Lerer and
                  James Bradbury and
                  Gregory Chanan and
                  Trevor Killeen and
                  Zeming Lin and
                  Natalia Gimelshein and
                  Luca Antiga and
                  Alban Desmaison and
                  Andreas K{\"{o}}pf and
                  Edward Z. Yang and
                  Zach DeVito and
                  Martin Raison and
                  Alykhan Tejani and
                  Sasank Chilamkurthy and
                  Benoit Steiner and
                  Lu Fang and
                  Junjie Bai and
                  Soumith Chintala},
  title        = {PyTorch: An Imperative Style, High-Performance Deep Learning Library},
  journal      = {CoRR},
  volume       = {abs/1912.01703},
  year         = {2019},
  eprinttype    = {arXiv},
  eprint       = {1912.01703},
  bibsource    = {dblp computer science bibliography, https://dblp.org}
}

@inproceedings{Zhang2023,
	author = {Dinghuai Zhang and Hanjun Dai and Nikolay Malkin and Aaron Courville and Yoshua Bengio and Ling Pan},
	booktitle = {Advances in Neural Information Processing Systems (NeurIPS)},
	title = {Let the Flows Tell: Solving Graph Combinatorial Optimization Problems with GFlowNets},
    year={2023},
}

@article{sachs,
  title={Causal protein-signaling networks derived from multiparameter single-cell data},
  author={Sachs, Karen and Perez, Omar and Pe'er, Dana and Lauffenburger, Douglas A and Nolan, Garry P},
  journal={Science},
  volume={308},
  number={5721},
  pages={523--529},
  year={2005},
  publisher={American Association for the Advancement of Science}
}

@inproceedings{eitl_uai2025,
author = {Ankan, Ankur and Textor, Johannes},
title = {Expert-in-the-loop causal discovery: iterative model refinement using expert knowledge},
year = {2025},
publisher = {JMLR.org},
booktitle = {Proceedings of the Forty-First Conference on Uncertainty in Artificial Intelligence},
articleno = {9},
numpages = {12},
location = {Rio de Janeiro, Brazil},
series = {UAI '25}
}

@inproceedings{llm_cd4,
title={Efficient Causal Graph Discovery Using Large Language Models},
author={Thomas Jiralerspong and Xiaoyin Chen and Yash More and Vedant Shah and Yoshua Bengio},
booktitle={ICLR 2024 Workshop: How Far Are We From AGI},
year={2024},
url={https://openreview.net/forum?id=5RBUTx75yr}
}

@article{llm_cd5,
author = {Vashishtha, Aniket and Reddy, Abbavaram Gowtham and Kumar, Abhinav and Bachu, Saketh and Balasubramanian, Vineeth N. and Sharma, Amit},
title = {Causal Inference Using LLM-Guided Discovery},
year = {2023},
month = {October},
url = {https://www.microsoft.com/en-us/research/publication/causal-inference-using-llm-guided-discovery/},
journal = {CoRR},
}

@misc{junyiLLM2025,
      title={Can Large Language Models Help Experimental Design for Causal Discovery?}, 
      author={Junyi Li and Yongqiang Chen and Chenxi Liu and Qianyi Cai and Tongliang Liu and Bo Han and Kun Zhang and Hui Xiong},
      year={2025},
      eprint={2503.01139},
      archivePrefix={arXiv},
      primaryClass={cs.AI},
      url={https://arxiv.org/abs/2503.01139}, 
}

@inproceedings{llm_uncert,
title={Can {LLM}s Express Their Uncertainty? An Empirical Evaluation of Confidence Elicitation in {LLM}s},
author={Miao Xiong and Zhiyuan Hu and Xinyang Lu and YIFEI LI and Jie Fu and Junxian He and Bryan Hooi},
booktitle={The Twelfth International Conference on Learning Representations},
year={2024},
url={https://openreview.net/forum?id=gjeQKFxFpZ}
}

@misc{llm_uncert_logits,
      title={Estimating LLM Uncertainty with Evidence}, 
      author={Huan Ma and Jingdong Chen and Joey Tianyi Zhou and Guangyu Wang and Changqing Zhang},
      year={2025},
      eprint={2502.00290},
      archivePrefix={arXiv},
      primaryClass={cs.CL},
      url={https://arxiv.org/abs/2502.00290}, 
}

@inproceedings{llm_quant_uncert,
  title={Quantifying uncertainty in natural language explanations of large language models},
  author={Tanneru, Sree Harsha and Agarwal, Chirag and Lakkaraju, Himabindu},
  booktitle={International Conference on Artificial Intelligence and Statistics},
  pages={1072--1080},
  year={2024},
  organization={PMLR}
}

@article{ribeiro2024anchorfci,
  title={AnchorFCI: harnessing genetic anchors for enhanced causal discovery of cardiometabolic disease pathways},
  author={Ribeiro, Adele H and Crnkovic, Milena and Pereira, Jaqueline Lopes and Fisberg, Regina Mara and Sarti, Flavia Mori and Rogero, Marcelo Macedo and Heider, Dominik and Cerqueira, Andressa},
  journal={Frontiers in Genetics},
  volume={15},
  pages={1436947},
  year={2024},
  publisher={Frontiers Media SA}
}

@ARTICLE{llm_cd1,
  author={Ban, Taiyu and Chen, Lyuzhou and Lyu, Derui and Wang, Xiangyu and Zhu, Qinrui and Chen, Huanhuan},
  journal={IEEE Transactions on Knowledge and Data Engineering}, 
  title={LLM-Driven Causal Discovery via Harmonized Prior}, 
  year={2025},
  volume={37},
  number={4},
  pages={1943-1960},
  doi={10.1109/TKDE.2025.3528461}
}

@inproceedings{dspy2024,
  title={DSPy: Compiling Declarative Language Model Calls into Self-Improving Pipelines},
  author={Khattab, Omar and Singhvi, Arnav and Maheshwari, Paridhi and Zhang, Zhiyuan and Santhanam, Keshav and Vardhamanan, Sri and Haq, Saiful and Sharma, Ashutosh and Joshi, Thomas T. and Moazam, Hanna and Miller, Heather and Zaharia, Matei and Potts, Christopher},
  journal={The Twelfth International Conference on Learning Representations},
  year={2024}
}

@article{dspy2022,
  title={Demonstrate-Search-Predict: Composing Retrieval and Language Models for Knowledge-Intensive {NLP}},
  author={Khattab, Omar and Santhanam, Keshav and Li, Xiang Lisa and Hall, David and Liang, Percy and Potts, Christopher and Zaharia, Matei},
  journal={arXiv preprint arXiv:2212.14024},
  year={2022}
}

@misc{kingma2017adammethodstochasticoptimization,
      title={Adam: A Method for Stochastic Optimization}, 
      author={Diederik P. Kingma and Jimmy Ba},
      year={2017},
      eprint={1412.6980},
      archivePrefix={arXiv},
      primaryClass={cs.LG},
      url={https://arxiv.org/abs/1412.6980}, 
}

@misc{zhang2024calibratingconfidencelargelanguage,
      title={Calibrating the Confidence of Large Language Models by Eliciting Fidelity}, 
      author={Mozhi Zhang and Mianqiu Huang and Rundong Shi and Linsen Guo and Chong Peng and Peng Yan and Yaqian Zhou and Xipeng Qiu},
      year={2024},
      eprint={2404.02655},
      archivePrefix={arXiv},
      primaryClass={cs.CL},
      url={https://arxiv.org/abs/2404.02655}, 
}

@misc{yang2024verbalizedconfidencescoresllms,
      title={On Verbalized Confidence Scores for LLMs}, 
      author={Daniel Yang and Yao-Hung Hubert Tsai and Makoto Yamada},
      year={2024},
      eprint={2412.14737},
      archivePrefix={arXiv},
      primaryClass={cs.CL},
      url={https://arxiv.org/abs/2412.14737}, 
}

@inproceedings{
    ban2025differentiable,
    title={Differentiable Structure Learning with Ancestral Constraints},
    author={Taiyu Ban and Changxin Rong and Xiangyu Wang and Lyuzhou Chen and Xin Wang and Derui Lyu and Qinrui Zhu and Huanhuan Chen},
    booktitle={Forty-second International Conference on Machine Learning},
    year={2025},
    url={https://openreview.net/forum?id=fyLsLTi1rE}
}

@article{Amershi2014,
  title = {Power to the People: The Role of Humans in Interactive Machine Learning},
  volume = {35},
  ISSN = {2371-9621},
  url = {http://dx.doi.org/10.1609/aimag.v35i4.2513},
  DOI = {10.1609/aimag.v35i4.2513},
  number = {4},
  journal = {AI Magazine},
  publisher = {Wiley},
  author = {Amershi,  Saleema and Cakmak,  Maya and Knox,  W. Bradley and Kulesza,  Todd},
  year = {2014},
  month = dec,
  pages = {105–120}
}

@misc{deleu2024discreteprobabilisticinferencecontrol,
      title={Discrete Probabilistic Inference as Control in Multi-path Environments}, 
      author={Tristan Deleu and Padideh Nouri and Nikolay Malkin and Doina Precup and Yoshua Bengio},
      year={2024},
      eprint={2402.10309},
      archivePrefix={arXiv},
      primaryClass={cs.LG},
      url={https://arxiv.org/abs/2402.10309}, 
}

@book{robert2007bayesian,
  title={The Bayesian choice: from decision-theoretic foundations to computational implementation},
  author={Robert, Christian P},
  year={2007},
  publisher={Springer}
}

@misc{tiapkin2024generativeflownetworksentropyregularized,
      title={Generative Flow Networks as Entropy-Regularized RL}, 
      author={Daniil Tiapkin and Nikita Morozov and Alexey Naumov and Dmitry Vetrov},
      year={2024},
      eprint={2310.12934},
      archivePrefix={arXiv},
      primaryClass={cs.LG},
      url={https://arxiv.org/abs/2310.12934}, 
}

@misc{malkin2023gflownetsvariationalinference,
      title={GFlowNets and variational inference}, 
      author={Nikolay Malkin and Salem Lahlou and Tristan Deleu and Xu Ji and Edward Hu and Katie Everett and Dinghuai Zhang and Yoshua Bengio},
      year={2023},
      eprint={2210.00580},
      archivePrefix={arXiv},
      primaryClass={cs.LG},
      url={https://arxiv.org/abs/2210.00580}, 
}

@misc{kipf2017semisupervisedclassificationgraphconvolutional,
      title={Semi-Supervised Classification with Graph Convolutional Networks}, 
      author={Thomas N. Kipf and Max Welling},
      year={2017},
      eprint={1609.02907},
      archivePrefix={arXiv},
      primaryClass={cs.LG},
      url={https://arxiv.org/abs/1609.02907}, 
}

@Article{pcalga,
    title = {Causal Inference Using Graphical Models with the {R}
      Package {pcalg}},
    author = {{Markus Kalisch} and {Martin M\"achler} and {Diego
      Colombo} and {Marloes H. Maathuis} and {Peter B\"uhlmann}},
    journal = {Journal of Statistical Software},
    year = {2012},
    volume = {47},
    number = {11},
    pages = {1--26},
    doi = {10.18637/jss.v047.i11},
}

@Article{pcalgaa,
    title = {Characterization and greedy learning of interventional
      {M}arkov equivalence classes of directed acyclic graphs},
    author = {{Alain Hauser} and {Peter B\"uhlmann}},
    journal = {Journal of Machine Learning Research},
    year = {2012},
    volume = {13},
    pages = {2409--2464},
    url = {https://jmlr.org/papers/v13/hauser12a.html},
}

@article{cohn1994active,
  title={Active learning with statistical models},
  author={Cohn, David and Ghahramani, Zoubin and Jordan, Michael},
  journal={Advances in neural information processing systems},
  volume={7},
  year={1994}
}

@book{peters2017elements,
  title={Elements of causal inference: foundations and learning algorithms},
  author={Peters, Jonas and Janzing, Dominik and Sch{\"o}lkopf, Bernhard},
  year={2017},
  publisher={The MIT press}
}

@misc{maddison2017concretedistributioncontinuousrelaxation,
      title={The Concrete Distribution: A Continuous Relaxation of Discrete Random Variables}, 
      author={Chris J. Maddison and Andriy Mnih and Yee Whye Teh},
      year={2017},
      eprint={1611.00712},
      archivePrefix={arXiv},
      primaryClass={cs.LG},
      url={https://arxiv.org/abs/1611.00712}, 
}

@misc{jang2017categoricalreparameterizationgumbelsoftmax,
      title={Categorical Reparameterization with Gumbel-Softmax}, 
      author={Eric Jang and Shixiang Gu and Ben Poole},
      year={2017},
      eprint={1611.01144},
      archivePrefix={arXiv},
      primaryClass={stat.ML},
      url={https://arxiv.org/abs/1611.01144}, 
}

@article{neal,
  author = {Neal, Radford M.},
  journal = {Handbook of Markov Chain Monte Carlo},
  pages = {113--162},
  title = {{MCMC} Using {Hamiltonian} Dynamics},
  volume = 54,
  year = 2010
}

@article{atanackovic2023dyngfn,
  title={DynGFN: Towards Bayesian Inference of Gene Regulatory Networks with GFlowNets},
  author={Atanackovic, Lazar and Tong, Alexander and Wang, Bo and Lee, Leo J and Bengio, Yoshua and Hartford, Jason S},
  journal={Advances in Neural Information Processing Systems},
  volume={36},
  year={2023}
}

@misc{viviano2025torchgfnpytorchgflownetlibrary,
      title={torchgfn: A PyTorch GFlowNet library}, 
      author={Joseph D. Viviano and Omar G. Younis and Sanghyeok Choi and Victor Schmidt and Yoshua Bengio and Salem Lahlou},
      year={2025},
      eprint={2305.14594},
      archivePrefix={arXiv},
      primaryClass={cs.LG},
      url={https://arxiv.org/abs/2305.14594}, 
}

@misc{paszke2019pytorchimperativestylehighperformance,
      title={PyTorch: An Imperative Style, High-Performance Deep Learning Library}, 
      author={Adam Paszke and Sam Gross and Francisco Massa and Adam Lerer and James Bradbury and Gregory Chanan and Trevor Killeen and Zeming Lin and Natalia Gimelshein and Luca Antiga and Alban Desmaison and Andreas Köpf and Edward Yang and Zach DeVito and Martin Raison and Alykhan Tejani and Sasank Chilamkurthy and Benoit Steiner and Lu Fang and Junjie Bai and Soumith Chintala},
      year={2019},
      eprint={1912.01703},
      archivePrefix={arXiv},
      primaryClass={cs.LG},
      url={https://arxiv.org/abs/1912.01703}, 
}

@software{jax2018github,
  author = {James Bradbury and Roy Frostig and Peter Hawkins and Matthew James Johnson and Chris Leary and Dougal Maclaurin and George Necula and Adam Paszke and Jake Vander{P}las and Skye Wanderman-{M}ilne and Qiao Zhang},
  title = {{JAX}: composable transformations of {P}ython+{N}um{P}y programs},
  url = {http://github.com/jax-ml/jax},
  version = {0.3.13},
  year = {2018},
}

@article{bollobas1980probabilistic,
  title={A probabilistic proof of an asymptotic formula for the number of labelled regular graphs},
  author={Bollob{\'a}s, B{\'e}la},
  journal={European Journal of Combinatorics},
  volume={1},
  number={4},
  pages={311--316},
  year={1980},
  publisher={Elsevier}
}

@book{newman2010networks,
  title={Networks: An Introduction},
  author={Newman, M.},
  isbn={9780191500701},
  url={https://books.google.ae/books?id=LrFaU4XCsUoC},
  year={2010},
  publisher={OUP Oxford}
}

@misc{foygel2010extendedbayesianinformationcriteria,
      title={Extended Bayesian Information Criteria for Gaussian Graphical Models}, 
      author={Rina Foygel and Mathias Drton},
      year={2010},
      eprint={1011.6640},
      archivePrefix={arXiv},
      primaryClass={math.ST},
      url={https://arxiv.org/abs/1011.6640}, 
}

@misc{drton2012iterativeconditionalfittinggaussian,
      title={Iterative Conditional Fitting for Gaussian Ancestral Graph Models}, 
      author={Mathias Drton and Thomas S. Richardson},
      year={2012},
      eprint={1207.4118},
      archivePrefix={arXiv},
      primaryClass={stat.ME},
      url={https://arxiv.org/abs/1207.4118}, 
}

@InProceedings{lagrange2025efficient,
  title = 	 {An Efficient Search-and-Score Algorithm for Ancestral Graphs using Multivariate Information Scores for Complex Non-linear and Categorical Data},
  author =       {Lagrange, Nikita and Isambert, Herve},
  booktitle = 	 {Proceedings of the 42nd International Conference on Machine Learning},
  pages = 	 {32164--32187},
  year = 	 {2025},
  editor = 	 {Singh, Aarti and Fazel, Maryam and Hsu, Daniel and Lacoste-Julien, Simon and Berkenkamp, Felix and Maharaj, Tegan and Wagstaff, Kiri and Zhu, Jerry},
  volume = 	 {267},
  series = 	 {Proceedings of Machine Learning Research},
  month = 	 {13--19 Jul},
  publisher =    {PMLR},
  url = 	 {https://proceedings.mlr.press/v267/lagrange25a.html},
}

@misc{ramsey2025scalablecausaldiscoveryrecursive,
      title={Scalable Causal Discovery from Recursive Nonlinear Data via Truncated Basis Function Scores and Tests}, 
      author={Joseph Ramsey and Bryan Andrews and Peter Spirtes},
      year={2025},
      eprint={2510.04276},
      archivePrefix={arXiv},
      primaryClass={stat.ML},
      url={https://arxiv.org/abs/2510.04276}, 
}

@article{madar2010dream3,
  title={DREAM3: network inference using dynamic context likelihood of relatedness and the inferelator},
  author={Madar, Aviv and Greenfield, Alex and Vanden-Eijnden, Eric and Bonneau, Richard},
  journal={PLoS ONE},
  year={2010},
  publisher={Public Library of Science},
}

@article{kloek1978bayesian,
  title={Bayesian estimates of equation system parameters: an application of integration by Monte Carlo},
  author={Kloek, Teun and Van Dijk, Herman K},
  journal={Econometrica: Journal of the Econometric Society},
  pages={1--19},
  year={1978},
  publisher={JSTOR}
}

@misc{openai2024gpt4technicalreport,
      title={GPT-4 Technical Report}, 
      author={OpenAI},
      year={2024},
      eprint={2303.08774},
      archivePrefix={arXiv},
      primaryClass={cs.CL},
      url={https://arxiv.org/abs/2303.08774}, 
}

@article{hassan2025efficient,
  title={Efficient autoregressive inference for transformer probabilistic models},
  author={Hassan, Conor and Loka, Nasrulloh and Li, Cen-You and Huang, Daolang and Chang, Paul E and Yang, Yang and Silvestrin, Francesco and Kaski, Samuel and Acerbi, Luigi},
  journal={arXiv preprint arXiv:2510.09477},
  year={2025}
}

@misc{chang2025amortizedprobabilisticconditioningoptimization,
      title={Amortized Probabilistic Conditioning for Optimization, Simulation and Inference}, 
      author={Paul E. Chang and Nasrulloh Loka and Daolang Huang and Ulpu Remes and Samuel Kaski and Luigi Acerbi},
      year={2025},
      eprint={2410.15320},
      archivePrefix={arXiv},
      primaryClass={stat.ML},
      url={https://arxiv.org/abs/2410.15320}, 
}

\newpage
\appendix

\renewcommand{\thesection}{\Alph{section}}

\section{Proofs} \label{sec:app:pa}

We provide rigorous proofs for our main statements in this section.

\subsection{Proof of Lemma~\ref{lemma:cycles}}

By definition, $\mathbf{F}_{ij} < \infty$ if and only if the distance from $V_{i}$ to $V_{j}$ is finite, i.e., there is a directed path from $V_{i}$ to $V_{j}$ (recall that the FW algorithm is solely applied on the directed component of the AG).

On the other hand, assume there is an almost directed path from $V_{i}$ to $V_{j}$.
Then, by definition, there are indices $k, l$ such that $V_{i} \rightsquigarrow V_{k}$ and $V_{k} \leftrightarrow V_{l}$ and $V_{l} \rightsquigarrow V_{j}$.
As explained, this implies $\mathbf{F}_{ik} < \infty$ and $\mathbf{F}_{lj} < \infty$ and $\tilde{\mathbf{B}}_{kl} < \infty$ (by definition).
Consequently, $\mathbf{T}_{iklj} \coloneqq \mathbf{F}_{ik} + \tilde{\mathbf{B}}_{kl} + \mathbf{F}_{lj} < \infty$, and $\min_{u, v} \mathbf{T}_{iuvj} \le \mathbf{T}_{iklj} < \infty$.

Conversely, if $\min_{u, v} \mathbf{T}_{iuvj} < \infty$, there are indices $k, l$ for which $\mathbf{T}_{iklj} < \infty$.
As a consequence, $\mathbf{F}_{ik} < \infty$ and $\tilde{\mathbf{B}}_{kl} < \infty$ $\mathbf{F}_{lj} < \infty$.
This implies there is a directed path from $V_{i}$ to $V_{k}$, $V_{k} \leftrightarrow V_{l}$, and a directed path from $V_{l}$ to $V_{j}$, i.e,. there is an almost directed path from $V_{i}$ to $V_{j}$.
\looseness=-1

We have shown that $\mathbf{F}$ and $\mathbf{T}$ are sufficient for detecting both directed and almost directed paths in an AG.
Importantly, both the FW algorithm and the tensor computation for $\mathbf{T}$ can be efficiently implemented in GPU.


\subsection{Proof of Proposition~\ref{prop:agfnsa}}

Our proof has two parts.
First, we argue that each AG compatible with the initial partial assignment can be generated through the iterative process described in \Cref{alg:onlinea}.
Second, we prove that the generative process only generates AGs.

Towards this objective, we 
initially show that \Cref{alg:onlinea} is equivalent to Lemma~\ref{lemma:cycles}.

\begin{lemma}[\Cref{alg:onlinea} is equivalent to Lemma~\ref{lemma:cycles}] \label{lemma:app:cycles}
    The directed path mask, $\boldsymbol{\alpha}$, in \Cref{alg:onlinea} is equivalent to $[\mathbf{F} < \infty]$ in Lemma~\ref{lemma:cycles}.
    Similarly, $\boldsymbol{\beta}$ is the same as $[\min_{u, v} \mathbf{T} < \infty]$ \looseness=-1
\end{lemma}

\begin{proof}
    By definition, the pair-wise distance matrix $\mathbf{F}$ satisfies $\mathbf{F}_{ij} < \infty$ for indices $i, j$ if and only if there is a directed path from $V_{i}$ to $V_{j}$ in the corresponding graph.
    We will show that $\boldsymbol{\alpha}_{ij} = 1$ if and only if there is a path from $V_{i}$ to $V_{j}$.

    Towards this objective, we proceed by induction on the set $P_{L} = \{(V_{i}, V_{j}) \colon d(V_{i}, V_{j}) \le L\}$, in which $d$ denotes the shortest distance on the graph.
    Assume $(V_{i}, V_{j}) \in P_{L}$ implies $\boldsymbol{\alpha}_{ij} = 1$.
    This is clearly true for $L = 1$.
    If $(V_{i}, V_{j}) \in P_{L + 1}$, then there are indices $k$, $l$ such that $(V_{k}, V_{l}) \in P_{1}$ and $(V_{i}, V_{k}) \in P_{I}$ and $(V_{l}, V_{j}) \in P_{J}$ for certain $I, J \le L$.
    Also, $(V_{k}, V_{l})$ were added to the graph after the paths $V_{i} \rightsquigarrow V_{k}$ and $V_{l} \rightsquigarrow V_{j}$ have already been formed.
    (These indices should always exist due to the iterative nature of \Cref{alg:onlinea}).
    By strong induction, $\boldsymbol{\alpha}_{ik} = 1$ and $\boldsymbol{\alpha}_{lj} = 1$.
    By \Cref{alg:onlinea}, $\boldsymbol{\alpha}_{ij} \ge \boldsymbol{\alpha}_{ik} \cdot \boldsymbol{\alpha}_{lj}$ and $\boldsymbol{\alpha}_{ij} = 1$.

    Conversely, assume that $\boldsymbol{\alpha}_{ij} = 1$.
    We will show by induction on the algorithm's step $T$ that there is a path from $V_{i}$ to $V_{j}$.
    Clearly, for $T = 1$, $\boldsymbol{\alpha}_{ij} = 1$ if and only if either there is a directed path from $V_{i}$ to $V_{j}$ (based on Lemma~\ref{lemma:cycles}) or $(V_{i}, V_{j}) \in P_{1}$ was the first added relationship.
    Assume, for $t \le T$, that $\boldsymbol{\alpha}_{ij} = 1$ on the $t$-th step of the algorithm implies $V_{i} \rightsquigarrow V_{j}$.
    Then, by definition, $\boldsymbol{\alpha}_{ij} = 1$ on the $(T + 1)$ step implies that either an edge from $V_{i}$ to $V_{j}$ was included, or there are indices $k, l$ such that $\boldsymbol{\alpha}_{ik} = 1$ and $\boldsymbol{\alpha}_{lj} = 1$ on the $T$-th step of the algorithm and the edge $(V_{k}, V_{l}) \in P_{1}$.
    By hypothesis, $\boldsymbol{\alpha}_{ik} = 1$ implies there is a directed path from $V_{i}$ to $V_{k}$.
    Similarly, $\boldsymbol{\alpha}_{lj} = 1$ implies there is a directed path from $V_{l}$ to $V_{j}$.
    As a consequence, there is a directed path from $V_{i}$ to $V_{j}$ on the $(T + 1)$-th step of the algorithm.
    By induction, $\boldsymbol{\alpha}_{ij} = 1$ implies that there is a directed path from $V_{i}$ to $V_{j}$.

    Based on the same reasoning and on this result, we can show that $\boldsymbol{\beta}_{ij} = 1$ if and only if there is an almost directed path from $i$ to $j$ in the AG.
\end{proof}

We show by induction that, for each $k$-sized subset of the $n$-sized variable set, \Cref{alg:onlinea} can generate each AG over the $k$ variables, $\mathbf{V}_{k} \subseteq \mathbf{V}$.
Clearly, for $k = 2$, if the initial condition is trivial (i.e., $S_{o} = \emptyset$ in Definition~\ref{def:agfnsga}), \Cref{alg:onlinea} can generate each AG over $\mathbf{V}_{k}$ by adding any relationships between the corresponding variable pair---resulting in graphs of the form $\mathcal{G}(f, S)$ for $S = \{\{V_{i}, V_{j}\}\}$ for any $\{V_{i}, V_{j}\} \in \mathbf{V}^{(2)}$ and any assignment function $f$; recall
Definition~\ref{def:agfnsga}.
If $S_{o} \neq \emptyset$, then there is either one or zero AG over $\mathbf{V}_{k}$ compatible with $S_{o}$, and our assertion follows trivially.

Assume that our generative process in \Cref{alg:onlinea} can construct every AG with $k - 1$ nodes abiding by any initial condition.
Consider the generative process over $k$-sized subsets of $\mathbf{V}$.
Let $\mathcal{G}$ be any AG over the $k$-sized variable set, and fix $V \in \mathcal{G}$ arbitrarily.

Let $E(V) = \{VrU \colon r \in \{\emptyset, \leftarrow, \rightarrow, \leftrightarrow\}, U \in \mathbf{V}_{k}\}$ be the set of ancestral relationships related to $V$.
Then, by definition, removing all relationships in $E(V)$ from $\mathcal{G}$ results in an AG $\mathcal{G}'$ over $\mathbf{V}_{k - 1}$, which is a $(k - 1)$-sized subset of $\mathbf{V}$. 
By the induction hypothesis, when restricted to $\mathbf{V}_{k - 1}$, our generative process can generate any AG, in particular, $\mathcal{G}'$.
By Lemmas~\ref{lemma:cycles} and~\ref{lemma:app:cycles}, each relationship that does not result in either a cycle or an almost cycle can be added to $\mathcal{G}'$.
As a consequence, our generative process can generate $\mathcal{G}$.
As $V$ and $\mathcal{G}$ were chosen arbitrarily, we have shown that our generative process can generate any AG over $k$-sized subsets of $\mathbf{V}$.
By induction, \Cref{alg:onlinea} is able to generate any AG with vertices on $\mathbf{V}$.

On the other hand, our generative process will only generate non-AGs if a transition introduces a cycle or almost cycle; however, by Lemma~\ref{lemma:app:cycles}, this cannot be the case.
As such, our generative process is restricted to the space of AGs, and generates all AGs compatible with a given initial condition.


\subsection{Proof of Lemma~\ref{lemma:posteriora}}

Recall $\mathbf{f}_{r}^{N} = (f_{r}^{(1)}, \dots, f_{r}^{(N)})$.
We henceforth omit the relation-related index $r$ for clarity.
Also, we let $\mathbf{f} = \mathbf{f}^{N}$.
For \Cref{eq:bayesian}, Bayes rule yields
\begin{equation*}
    q(\omega|\mathbf{f}) \propto p(\omega) \prod_{1 \le n \le N} p(f^{(n)} | \omega) = p(\omega) \prod_{1 \le n \le N} \pi^{[f^{(n)} = \omega]} \cdot \left( \frac{1 - \pi}{3} \right)^{[f^{(n)} \neq \omega]} = p(\omega) \pi^{n_{\omega}} \left( \frac{1 - \pi}{3} \right)^{N - n_{\omega}},
\end{equation*}
in which $n_{\omega} = \sum_{n=1}^{N} [f^{(n)} = \omega]$.
As $\omega \in \{\emptyset, \leftarrow, \rightarrow, \leftrightarrow\}$, we can easily normalize the right-hand side of the above equation.
This provides a formula for the posterior distribution when $\pi$ is fixed or given. When $\pi$ follows a Beta distribution with parameters $\alpha$ and $\beta$, we should marginalize the joint posterior over $\pi$.
That is,
\begin{equation*}
    q(\omega |\mathbf{f}) \propto \int_{0}^{1} p(\omega) p(\pi) \prod_{1 \le n \le N} p(f^{(n)} | \omega) \mathrm{d}\pi = \int_{0}^{1} p(\omega) \pi^{\alpha - 1} (1 - \pi)^{\beta - 1} \pi^{n_{\omega}} \left( \frac{1 - \pi}{3} \right)^{N - n_{\omega}} \mathrm{d}\pi.
\end{equation*}
Since
\begin{equation*}
    \begin{split}
        \int_{0}^{1}\pi^{\alpha - 1} (1 - \pi)^{\beta - 1} \pi^{n_{\omega}} \left( \frac{1 - \pi}{3} \right)^{N - n_{\omega}} \mathrm{d}\pi = \\ 3^{n_{\omega} - N} \int_{0}^{1} \pi^{n_{\omega} + \alpha - 1} (1 - \pi)^{N - n_{\omega} +  \beta - 1} \mathrm{d}\pi = 3^{n_{\omega} - N} B(n_{\omega} + \alpha, N - n_{\omega} + \beta), 
    \end{split}
\end{equation*}
in which $B(x, y) \coloneqq \nicefrac{\Gamma(x) \Gamma(y)}{\Gamma(x + y)}$ is the Beta function and $\Gamma$ is the Gamma function. Hence, \looseness=-1
\begin{equation*}
    q(\omega | \mathbf{f}) \propto 3^{n_{\omega} - N} \cdot \boldsymbol{\rho}^{(\omega)} \cdot B(n_{\omega} + \alpha, N - n_{\omega} + \beta),
\end{equation*}
in which $\boldsymbol{\rho}^{(\omega)}$ is the probability for $\omega$ according to our categorical prior (recall \Cref{eq:bayesian}).
As the Beta function can be efficiently and accurately computed, this ensures that both posteriors are semi-tractable. \looseness=-1

\subsection{Proof of Proposition~\ref{prop:consistencya}}

We will in this section denote $\pi^{\star}$ as a vector in $\mathbb{R}^{4}$ such that $\pi^{\star}_{\omega^{\star}} = \pi^{\star} \in \mathbb{R}$ and $\pi^{\star}_{\omega} = \nicefrac{1 - \pi^{\star}}{3}$ for $\omega^{\star} \neq \omega$.
Thus the condition $\pi^{\star} > \nicefrac{1}{4}$ simply means that 
$\pi^{\star}_{\omega^{\star}} > \pi_{\omega}^{\star}$ for each $\omega \in \{\emptyset, \leftarrow, \rightarrow, \leftrightarrow\}$ and $\omega \neq \omega^{\star}$.
Similarly to the previous section, we omit the relation index $r$ from our computations in favor of clarity.
Then, we will demonstrate that our posterior distribution is consistent when the model is identifiable and the expert's feedback is better-than-random.

Clearly, the case $\pi^{\star}_{\omega^{\star}} = 1$ is trivial; henceforth, assume that $\pi^{\star}_{\omega^{\star}} < 1$.
Intuitively, our proof will show that the log-ratio of $\frac{1}{N} \log \frac{q(\omega^{\star} | \mathbf{f})}{q(\omega | \mathbf{f})}$ is almost surely positive on $\mathbf{f}$ 
when $\omega \neq \omega^{\star}$ and $N \rightarrow \infty$.
This can only happen if $q(\omega^{\star} | \mathbf{f})$ grows exponentially faster than $q(\omega | \mathbf{f})$, which ensures the consistency of our posterior distribution.

We start with the model in \Cref{eq:bayesianpriors}, in which $\pi$ is not fixed, and proceed to \Cref{eq:bayesian} later.
Based on Lemma~\ref{lemma:posteriora},
\begin{equation*}
    \frac{1}{N} \log \frac{q(\omega^{\star} | \mathbf{f})}{q(\omega | \mathbf{f})} = \frac{1}{N} \left( (n_{\omega^{\star}} - n_{\omega}) \cdot (\log 3) + \log \frac{\boldsymbol{\rho}^{(\omega^{\star})}}{\boldsymbol{\rho}^{(\omega)}} + \log \frac{B(n_{\omega^{\star}} + \alpha, N - n_{\omega^{\star}} + \beta)}{B(n_{\omega} + \alpha, N - n_{\omega} + \beta)} \right).
\end{equation*}
By the definition of the Beta function,
\begin{equation} \label{eq:app:betas}
    \log \frac{B(n_{\omega^{\star}} + \alpha, N - n_{\omega^{\star}} + \beta)}{B(n_{\omega} + \alpha, N - n_{\omega} + \beta)} = \log \frac{\Gamma(n_{\omega^{\star}} + \alpha) \Gamma(N - n_{\omega^{\star}} + \beta)}{\Gamma(n_{\omega} + \alpha) \Gamma(N - n_{\omega} + \beta)} \cdot \frac{\Gamma(N + \alpha + \beta)}{\Gamma(N + \alpha + \beta)}.
\end{equation}
As we are only interested in the asymptotics, 
we approximate $\log \Gamma$ via 
Stirling's formula, \looseness=-1
\begin{equation*}
    \log \Gamma(x) = \frac{1}{2} \log (2 \pi) + \left(x - \frac{1}{2}\right) \log x - x + \mathcal{O}\left(\frac{1}{x}\right).
\end{equation*}
In particular, for $x, y > 0$,
\begin{equation*}
    \log \frac{\Gamma(x)}{\Gamma(y)} = \left( x - \frac{1}{2} \right) \log x - \left( y - \frac{1}{2} \right) \log y - (x - y)  + \mathcal{O}\left( \frac{1}{x} \right) - \mathcal{O}\left(\frac{1}{y}\right).
\end{equation*}
By plugging $x = n_{\omega} + \alpha$ and $y = N - n_{\omega} + \beta$ into the above equation, we obtain
\begin{equation*}
    \begin{split}
        \log \frac{\Gamma(n_{\omega} + \alpha)}{\Gamma(N - n_{\omega} + \beta)} =  \left( n_{\omega} + \alpha - \frac{1}{2} \right) \log (n_{\omega} + \alpha) - \left( N - n_{\omega} + \beta - \frac{1}{2} \right) \log (N - n_{\omega} + \beta) \\ - ((n_{\omega} + \alpha) - (N - n_{\omega} + \beta))  + \mathcal{O}\left( \frac{1}{n_{\omega} + \alpha} \right) - \mathcal{O}\left(\frac{1}{N - n_{\omega} + \beta}\right).
    \end{split}
\end{equation*}
Similarly, a Taylor approximation of $\log (n_{\omega} + \alpha)$ around $n_{\omega}$ yields
\begin{equation*}
    \log (n_{\omega} + \alpha) = \log n_{\omega} + \frac{\alpha}{n_{\omega}} + \mathcal{O}\left( \frac{1}{n_{\omega}} \right) \text{ and } \log (N - n_{\omega} + \beta) = \log (N - n_{\omega}) + \frac{\beta}{N - n_{\omega}}  + \mathcal{O}\left( \frac{1}{N - n_{\omega}} \right).
\end{equation*}
Hence,
\begin{equation*}
    \frac{n_{\omega}}{N} \log (n_{\omega} + \alpha) = \frac{n_{\omega}}{N}  \log n_{\omega} + \frac{\alpha}{N} + \mathcal{O}\left( \frac{1}{N} \right), 
\end{equation*}
and similarly for $N - n_{\omega} + \beta$.
Courage!
We may ignore most of these terms during an asymptotic computation.
To see this, we divide both members of our Stirling approximation by $N$,
\begin{equation*}
    \begin{split}
        \frac{1}{N} \log \frac{\Gamma(n_{\omega} + \alpha)}{\Gamma(N - n_{\omega} + \beta)} =  \frac{1}{N} \left( n_{\omega} + \alpha - \frac{1}{2} \right) \log (n_{\omega} + \alpha) - \frac{1}{N} \left( N - n_{\omega} + \beta - \frac{1}{2} \right) \log (N - n_{\omega} + \beta) \\ - \left( \frac{ 2 n_{\omega} - N + \alpha - \beta}{N} \right) + \mathcal{O}\left( \frac{1}{N( n_{\omega} + \alpha) } \right) - \mathcal{O}\left(\frac{1}{N(N - n_{\omega} + \beta)}\right).
    \end{split}
\end{equation*}
When $N \rightarrow \infty$, $n_{\omega} \rightarrow \infty$ and $N - n_{\omega} \rightarrow \infty$ almost surely (as $\nicefrac{n_{\omega}}{N} \rightarrow \pi_{\omega}^{\star} > 0$ and $\nicefrac{N - n_{\omega}}{N} \rightarrow 1 - \pi_{\omega}^{\star} > 0$ by assumption); hence, ignoring low-order terms in the equation above,
\begin{equation*}
    \begin{aligned}
        \frac{1}{N} \log \frac{\Gamma(n_{\omega} + \alpha)}{\Gamma(N - n_{\omega} + \beta)} &= \frac{n_{\omega}}{N} \log n_{\omega} - \frac{N - n_{\omega}}{N} \log (N - n_{\omega}) - \frac{2n_{\omega} - N}{N} + \mathcal{O}\left( \frac{\log N}{N} \right) \\
        &= \frac{n_{\omega}}{N} \log (n_{\omega} \cdot (N - n_{\omega})) - \log (N - n_{\omega}) - \frac{2n_{\omega} - N}{N} +  \mathcal{O}\left(\frac{\log N}{N}\right). 
    \end{aligned}
\end{equation*}
This equation also holds for $\omega^{\star}$ in place of $\omega$.
Going back to \Cref{eq:app:betas}, we observe that
\begin{equation*}
    \begin{aligned}
        \frac{1}{N} \log \frac{B(n_{\omega^{\star}} + \alpha, N - n_{\omega^{\star}} + \beta)}{B(n_{\omega} + \alpha, N - n_{\omega} + \beta)} &= \frac{1}{N} \log \frac{\Gamma(n_{\omega^{\star}} + \alpha) \Gamma(N - n_{\omega^{\star}} + \beta)}{\Gamma(n_{\omega} + \alpha) \Gamma(N - n_{\omega} + \beta)}
        \\
        &= \frac{n_{\omega^{\star}}}{N} \log ( n_{\omega^{\star}} \cdot (N - n_{\omega^{\star}})) - \frac{n_{\omega}}{N} \log  ( n_{\omega} \cdot (N - n_{\omega})) \\ &- \log \frac{N - n_{\omega^{\star}}}{N - n_{\omega}} + \frac{2 n_{\omega} - 2 n_{\omega^{\star}}}{N} + \mathcal{O}\left( \frac{\log N}{N} \right).
    \end{aligned}
\end{equation*}
By writing $n_{\omega} = N p_{\omega}$ for some $p_{\omega} \in [0, 1]$, then
\begin{equation*}
    \begin{aligned}
        \frac{1}{N} \log \frac{B(n_{\omega^{\star}} + \alpha, N - n_{\omega^{\star}} + \beta)}{B(n_{\omega} + \alpha, N - n_{\omega} + \beta)} &= \frac{n_{\omega^{\star}}}{N} \log ( n_{\omega^{\star}} \cdot (N - n_{\omega^{\star}})) - \frac{n_{\omega}}{N} \log  ( n_{\omega} \cdot (N - n_{\omega})) \\ &- \log \frac{N - n_{\omega^{\star}}}{N - n_{\omega}} + \frac{2 n_{\omega} - 2 n_{\omega^{\star}}}{N} + \mathcal{O}\left( \frac{\log N}{N} \right) \\
        &= p_{\omega^{\star}} \log ( p_{\omega^{\star}} \cdot (1 - p_{\omega^{\star}})) - p_{\omega} \log ( p_{\omega} \cdot (1 - p_{\omega})) \\
        &+ 2 (p_{\omega^{\star}} - p_{\omega}) \log N + 2 (p_{\omega} - p_{\omega^{\star}}) - \log \frac{1 - p_{\omega^{\star}}}{1 - p_{\omega}} + \mathcal{O}\left(\frac{\log N}{N}\right).
    \end{aligned}
\end{equation*}
As a consequence,
\begin{equation*}
    \begin{aligned}
        \frac{1}{N} \log \frac{q(\omega^{\star} | \mathbf{f})}{q(\omega | \mathbf{f})} &= \frac{1}{N} \left( (n_{\omega^{\star}} - n_{\omega}) \cdot (\log 3) + \log \frac{\boldsymbol{\rho}^{(\omega^{\star})}}{\boldsymbol{\rho}^{(\omega)}} + \log \frac{B(n_{\omega^{\star}} + \alpha, N - n_{\omega^{\star}} + \beta)}{B(n_{\omega} + \alpha, N - n_{\omega} + \beta)} \right) \\
        &= (p_{\omega^{\star}} - p_{\omega}) ( 2 \log N + \log 3 - 2) \\
        &+ p_{\omega^{\star}} \log ( p_{\omega^{\star}} \cdot (1 - p_{\omega^{\star}})) - p_{\omega} \log ( p_{\omega} \cdot (1 - p_{\omega}))
        \\ &+ \frac{1}{N} \log \frac{\boldsymbol{\rho}^{(\omega^{\star})}}{\boldsymbol{\rho}^{(\omega)}} - \log \frac{1 - p_{\omega^{\star}}}{1 - p_{\omega}} + \mathcal{O}\left(\frac{\log N}{N}\right).
    \end{aligned}
\end{equation*}
Obviously, $2\log N + \log 3 - 2 > 0$ when $N \ge 2$ and $p_{\omega^{\star}} > p_{\omega}$ almost surely for sufficiently large $N$ when the expert's feedback is better-than-random.
Also, the right-hand-side of the equation above is clearly dominated by $\log N$.
That is,
\begin{equation*}
    \begin{aligned}
        \frac{1}{N\log N} \log \frac{q(\omega^{\star} | \mathbf{f})}{q(\omega | \mathbf{f})} \rightarrow 2 (\pi_{\omega^{\star}} - \pi_{\omega} )
    \end{aligned}
\end{equation*}
when $N \rightarrow \infty$. Consequently, $q(\omega^{\star} | \mathbf{f})$ grows exponentially faster than $q(\omega | \mathbf{f})$, and their quotient diverges to $\infty$.
This can only happen if $q(\omega^{\star} |\mathbf{f}) \rightarrow 1$ as $N \rightarrow \infty$.
As such, we have shown the posterior concentration of our model on the true ancestral relationship of each relation $r$.

In this context, we presently show that our model with fixed $\pi$ also converges to a point mass in $\omega^{\star}$ when the expert's feedback is sufficiently accurate.
Initially, notice that the log-ratio for the posterior distributions are
\begin{equation*}
    \frac{1}{N} \log \frac{q(\omega^{\star}|\mathbf{f})}{q(\omega|\mathbf{f})} = \frac{1}{N} \log \frac{\boldsymbol{\rho}^{(\omega^{\star}})}{\boldsymbol{\rho}^{(\omega)}} + \frac{(n_{\omega^{\star}} - n_{\omega})}{N} \log \pi + \frac{(n_{\omega} - n_{\omega^{\star}})}{N} \log \frac{1 - \pi}{3}.
\end{equation*}
When $N \rightarrow \infty$, 
this converges almost surely to (due to the law of large numbers) \looseness=-1
\begin{equation} \label{eq:app:asymptotics}
    \begin{aligned}
        \frac{1}{N} \log \frac{q(\omega^{\star}|\mathbf{f})}{q(\omega|\mathbf{f})} &= \left( \pi - \frac{1 - \pi}{3} \right) \log \pi + \left( \frac{1 - \pi}{3} - \pi \right) \log \frac{1 - \pi}{3} \\
        &= \frac{4\pi - 1}{3} \log \pi + \frac{1 - 4 \pi}{3} \log \frac{1 - \pi}{3} \\
        &= \frac{4 \pi - 1}{3} \log \frac{3\pi}{1 - \pi}.
    \end{aligned}
\end{equation}
This quantity is positive when $\pi > \frac{1}{4}$.
Consequently, the posterior distribution also concentrates in the true relationship $\omega^{\star}$.
In conclusion, we have shown that both our models for the expert's feedback provide consistent estimates of the true AG.

\subsection{Proof of Proposition~\ref{prop:mconsistencya}}

We proceed similarly to the above section.
As before, we will drop the indices $r, N$ for clarity.
Then, under \Cref{eq:bayesian},
\begin{equation*}
    \frac{1}{N} \log \frac{q(\omega^{\star}|\mathbf{f})}{q(\omega|\mathbf{f})} = \frac{1}{N} \log \frac{\boldsymbol{\rho}^{(\omega^{\star})}}{\boldsymbol{\rho}^{(\omega)}} + \frac{n_{\omega^{\star}} - n_{\omega}}{N} \log \pi + \frac{n_{\omega} - n_{\omega^{\star}}}{N} \log \frac{1 - \pi}{3}.
\end{equation*}
By the law of large numbers, the ratio $\nicefrac{n_{\omega^{\star}}}{N}$ converges almost surely to $\pi^{\star}$, while $\nicefrac{n_{\omega}}{N} \rightarrow \nicefrac{1 - \pi^{\star}}{3}$ as $N \rightarrow \infty$.
Consequently, the ratio of log-posteriors above satisfies
\begin{equation*}
    \frac{1}{N} \log \frac{q(\omega^{\star} | \mathbf{f})}{q(\omega|\mathbf{f})} \rightarrow \frac{4 \pi^{\star} - 1}{3} \log \frac{3\pi}{1 - \pi} 
\end{equation*}
by the same reasoning of \Cref{eq:app:asymptotics}.
When $\pi > \nicefrac{1}{4}$ and $\pi^{\star} > \nicefrac{1}{4}$, the right-hand side of the above equation is positive.
Hence, $q(\omega^{\star}|\mathbf{f}) \rightarrow 1$ almost surely as $N \rightarrow \infty$.

\section{On the knowledge of (non-)ancestral relationships}
\label{sec:app:knownledge}

Our EITL approach is based on the premise that an expert can determine whether a variable is a cause (ancestor) of another, regardless of whether this causation is direct or confounded.
This knowledge is much coarser compared to that needed for a causal DAG. While directed edges in a causal DAG represent direct causation (assumed to be unconfounded), which can be challenging for humans to identify, edges in an AG represent only the existence of a directed (causal) path, regardless of any other potentially unobserved, mediators or confounding paths.

Scientists often have an understanding of the ancestral (or non-ancestral) relationship between two variables. Illustratively, causal knowledge primarily arises from randomized experiments. In biomedicine, for example, such information is readily available on platforms such as PubMed \citep{wheeler2007database}, DrugBank \citep{knox2024drugbank}, and GO-CAM \citep{thomas2019gene}. Yet, it is important to note that causality from experimental studies only guarantees ancestrality, not directness or unconfoundedness. For example, causal effects may be mediated by unobserved factors, and confounding, albeit reduced by randomization in experiments, can persist in observational studies. An expert can incorporate these established ancestral relations using our framework by increasing the probability of the corresponding directed edges.

There are also numerous examples of known non-ancestral relationships. For instance, it is widely accepted that sociodemographic variables (e.g., age and sex), as well as genetic variables (referred to as $X$), are not caused by factors such as drugs, diseases, or other phenotypes (referred to as $Y$). To reflect this understanding that $Y$ is not an ancestor of $X$, one can reduce the likelihood of $Y$ being ancestral to $X$, achieved by decreasing the probability of the corresponding edges $Y \rightarrow X$.

Additionally, information indicating that an association is purely spurious, with no causation in any direction, may also be accessible. For instance, the link between coffee consumption ($X$) and heart disease ($Y$) is likely to be solely spurious, as factors such as smoking, diet, and lifestyle habits among heavy coffee drinkers could confound the results. This knowledge is incorporated by increasing the probability of a bidirected edge $X \leftrightarrow Y$.

Finally, our EITL approach considers the confidence level of such knowledge, facilitating its inclusion even when the source may lack reliability.

\section{Experimental details} \label{sec:app:experimentsa}

We briefly overview our experimental setup here.
Readers may consult the accompanying code for further details on the implementation of AGFN and of our EITL algorithm. \looseness=-1

\subsection{Estimating the number of AGs}

To estimate the number of AGs in \Cref{fig:nodesa}, we pick a graph uniformly at random and verify, using \cite{bhattacharya2021differentiable}'s algebraic characterization of ancestral graphs, whether it satisfies the ancestrality constraints.
Then, the estimated number of AGs is simply the total number of graphs multiplied by the estimated proportion.
In more details, there are $4^{\binom{n}{2}}$ graphs with $n$ nodes and edges in $\{\emptyset, \leftarrow, \rightarrow, \leftrightarrow\}$.
Hence, we sample $K$ graphs, $\{\mathcal{G}_{k}\}_{k=1}^{K}$, and compute
\begin{equation*}
    \delta_{K} = \left( \frac{1}{K} \sum_{1 \le k \le K} \mathrm{isAncestral}(\mathcal{G}_{k}) \right) \cdot 4^{\binom{n}{2}}, 
\end{equation*}
in which $\mathrm{isAncestral}$ is defined as
\begin{equation*}
    \mathrm{isAncestral}(\mathcal{G}) = [ \mathtt{tr} \left( \exp\{\mathbf{D}\} \right) - n + \mathtt{sum} ( \exp\{\mathbf{D}\} \odot \mathbf{B})  = 0],
\end{equation*}
with $\mathtt{tr}$ as the trace operator.
Obviously, $\delta_{K}$ is an unbiased and consistent estimator of the number of AGs with a given number of nodes.
We let $K = 5 \cdot 10^{7}$ for all experiments in \Cref{fig:nodesa}.
To avoid floating point overflow, we compute $4^{\binom{n}{2}}$ and $\delta_{K}$ in the log-space. \looseness=-1

\subsection{Posterior concentration}

The plot in \Cref{fig:posteriora} was obtained by simulating the models in \Cref{eq:bayesian,eq:bayesianpriors} with given parameters $\omega^{\star}$ and $\pi_{r}^{\star} = 0.9$ and increasing dataset sizes.
For the misspecified model, we let $\pi_{r} = 0.6$. 
For our fully Bayesian model in \Cref{eq:bayesianpriors}, we use an uniform prior over $\pi_{r}$.
Importantly, notice that our observations are in agreement with Propositions~\ref{prop:consistencya} and~\ref{prop:mconsistencya}.
Results and error bars were computed across 100 simulations.

\subsection{AGFN and EITL pipeline}

This subsection delineates the setup for our experimental campaign in \Cref{sec:experiments}.

\subsubsection{Baselines}

\paragraph{FCI} We first estimated a PAG using the stable version of FCI, which produces a fully order-independent final skeleton \citep{colombo2014order}. To identify conditional independencies, we used Fisher's Z partial correlation test with a significance level of $\alpha = 0.05$. The BIC score associated with the PAG estimated by the FCI was computed as the BIC of a randomly selected maximal AG (MAG) within the equivalence class characterized by such PAG. The maximality of an AG depends on the absence of inducing paths between non-adjacent variables, which are paths where every node along it (except the endpoints) is a collider and every collider is an ancestor of an endpoint \citep{DBLP:conf/uai/RantanenHJ21}. This ensures that in the MAG every non-adjacent pair of nodes is m-separated by some set of other variables. Importantly, Markov equivalent MAGs exhibit asymptotic equivalence in terms of BIC scores \citep{richardson2002ancestral}. As a result, the choice of a random MAG does not disrupt the validity of our results.

\paragraph{GFCI} Similarly, we applied GFCI with an initial search algorithm (FGS) based on the BIC score and the subsequent application of the FCI with conditional independencies identified by the Fisher's Z partial correlation test with a significance level $\alpha = 0.05$.
Also, similarly to FCI,
the BIC score associated with the estimated PAG was computed as the BIC of a randomly selected MAG within the equivalence class characterized by such PAG.
\looseness=-1

\paragraph{ACI} We used the implementation provided by the authors at GitHub\footnote{ACI repository, available online at  \href{https://github.com/caus-am/aci}{https://github.com/caus-am/aci}}. For the Answer Set Programming solver, we used clingo 4, version 5.6.2, also available at GitHub \footnote{Clingo repository, available online at  \href{https://github.com/potassco/clingo}{https://github.com/potassco/clingo}.}. Similarly to the FCI, we tested conditional independencies using Fisher's Z partial correlation test with a significance level of $\alpha = 0.05$.  ACI also outputs a PAG, so we computed the BIC of a randomly selected MAG within the represented Markov equivalence class.

\paragraph{DCD} We adhered to the instructions provided in the official repository\footnote{Available online at \href{https://gitlab.com/rbhatta8/dcd}{https://gitlab.com/rbhatta8/dcd}.} to apply the DCD method. 
The SHD was obtained between the ground-truth PAG and the PAG corresponding to the estimated ADMG (i.e., the one obtained via FCI by using the d-separations entailed by the estimated ADMG as an oracle for conditional independencies).
On the other hand, the BIC was computed for the estimated ADMG directly.

\paragraph{N-ADMG} To estimate the parameters of the variational distribution defined by N-ADMG, we executed the code provided at the official repository\footnote{Available online at \href{https://github.com/microsoft/causica/releases/tag/v0.0.0}{https://github.com/microsoft/causica/releases/tag/v0.0.0}.}  For fairness, we used the same hyperparameters and architectures reported in their original work \citep{DBLP:conf/iclr/LiLSH23}; in particular, we trained the models for $30$k epochs. After this, we sampled $100$k graphs from the learned distribution. It is worth mentioning that the constraints of bow-free ADMG are guaranteed in the N-ADMG samples only in an asymptotic sense. Thus, we manually removed any cyclic graphs from the learned distribution. Then, we proceeded exactly as with DCD to estimate both the average SHD and the average BIC under the variational distribution.

Additionally, we also considered GPS\footnote{Available online at \href{https://github.com/tomc-ghub/gps_uai2022}{https://github.com/tomc-ghub/gps\_uai2022}.} \cite{claassen2022greedy}, however, we noticed its runtime was orders of magnitude larger than alternative approaches.
As a consequence, we excluded GPS from our main experimental campaign.

\subsubsection{Sachs dataset details}\label{sec:app:sachs}
Sachs~\citep{sachs} is a dataset that measures the levels of specific proteins and phospholipids in the human cell. The data are continuous and the $11$ observed variables/features are: \texttt{"Raf", "Mek", "Plcg", "PIP2", "PIP3", "Erk", "Akt", "PKA", "PKC", "P38", "Jnk"}.

As we would like the dataset to contain latent counfounders, we used three different versions in our experiments: \texttt{Sachs}, \texttt{Sachs-7} and \texttt{Sachs-5}. All have an equal number of samples, $7466$---which is the original amount.

\texttt{Sachs} is the original dataset. In order to reduce the number of variables, we apply a simple conditional independence test and remove the ones with the most false independencies. Thus, we ensure the resulting graphs remain sufficiently connected and not overly sparse.

\texttt{Sachs-7} contains the following variables: \texttt{"Mek", "Plcg", "PIP2", "PIP3", "PKA", "PKC", "Jnk"} and \texttt{Sachs-5}: \texttt{"Mek", "PIP3", "PKA", "PKC", "Jnk"}.

\subsection{Details on the Experiments with LLMs}\label{sec:app:llm_exp}

Our LLM-based framework is built using DSPy~\citep{dspy2022,dspy2024}, a library to build modular AI software. We define to the LLM what the input and output will be:

\begin{python}
query: tuple[str, str] = dspy.InputField(
    description=(
        "Pair of nodes about which we want to know the relationship. "
        "Each pair of nodes may be directly associated, via direct causation, or indirectly associated, via latent confounding. "
        "The nature of each node and the underlying problem is described in the context field."
    )
)

context: str = dspy.InputField(
    description=(
        "The context of the problem, including the nature of each node (represented by a string) "
        "and the nature of the kind of relationships we are looking for."
    )
)

relationship: EdgeType = dspy.OutputField(
    description=(
        "This is the feedback. It describes the nature of the relationship between the input nodes. "
        "This information should be inferred from the provided context, background knowledge, and the nature of the problem."
    )
)
\end{python}

Then, for each given query, we prompt the LLM:

\begin{tcolorbox}[
    colback=yellow!10,
    colframe=black!40,
    boxrule=0.5pt,
    arc=2mm,
    left=4pt,
    right=4pt,
    top=4pt,
    bottom=4pt,
]
You are an expert on the human immune system cell.
You are investigating the cause-and-effect relationships between a specific
set of observed variables representing proteins and phospholipids:
"Raf", "Mek", "Plcg", "PIP2", "PIP3", "Erk", "Akt", "PKA", "PKC", "P38", "Jnk".
Your task is to determine the ancestral relationship between the given variables.
If there are both indirect and direct relationships between the variables,
you should describe only the direct one.
\end{tcolorbox}

\paragraph{Uncertainty over LLM feedback.}
Quantifying the uncertainty of LLM responses remains an active area of research \citep{llm_quant_uncert}. While explicitly prompting the model to self-report confidence is one approach, it has proven to be unreliable \citep{llm_uncert}. Alternatively, other methods rely on analyzing output logits to estimate uncertainty \citep{llm_uncert_logits}. Unfortunately, this technique is inapplicable to state-of-the-art closed-source models, such as GPT-4o, where access to logits is restricted. For this reason, we use a Monte Carlo estimate with $10$ samples for each query.

\subsubsection{Training hyperparameters}

For AGFN's forward policy, we use a MLP 
with $2$ layers to compute embeddings of dimension $256$ interleaved by leaky ReLU layers.
We fix the backward flow as uniform.
For training, we use the Adam method for the stochastic optimization problem defined by the minimization of the TB loss. Moreover, we trained the neural networks for $3000$ epochs for the human-in-the-loop simulations (in which we considered graphs having up to $10$ nodes) and for $500$ epochs for both the assessment of the distributional quality of AGFN and the comparison of AGFN with alternative CD approaches in \Cref{fig:agfnsdists}.

\subsubsection{Computational settings}\label{sec:app:comp_sett}
We trained the AGFNs in \Cref{sec:experiments} computer clusters equipped with NVIDIA's V100 GPUs and on a local workstation featuring a NVIDIA RTX 4070 GPU.
The algorithms were implemented using the machine learning framework \texttt{PyTorch} \citep{torch}. To estimate the PAG corresponding to AGFN's samples and compute the SHDs, we used the FCI's implementation of the \texttt{pcalg} package \citep{pcalga, pcalgaa} in \texttt{R} considering the d-separations entailed by these samples as a criterion for conditional dependence.

\end{document}